\documentclass{article}
\PassOptionsToPackage{table}{xcolor}
\usepackage{iclr2027/iclr2027_conference,times}

\usepackage[utf8]{inputenc} 
\usepackage[T1]{fontenc}    
\usepackage{hyperref}       
\hypersetup{linkbordercolor=cyan, citebordercolor=cyan, urlbordercolor=cyan}
\usepackage{amsmath,amsfonts,bm}
\usepackage[capitalise]{cleveref}       
\usepackage{url}            
\usepackage{booktabs}       
\usepackage{amsfonts}       
\usepackage{nicefrac}       
\usepackage{microtype}      
\usepackage{graphicx}
\usepackage{xspace}
\usepackage{enumitem}
\usepackage{multirow}
\usepackage[table]{xcolor} 
\usepackage{booktabs}      
\usepackage{tabularx}
\usepackage{pifont}       
\usepackage{placeins}     
\usepackage{pgfplots}
\pgfplotsset{compat=1.18}
\usepgfplotslibrary{groupplots}
\usetikzlibrary{calc}
\definecolor{jttXS}{HTML}{2A78D6}
\definecolor{jttS}{HTML}{EB6834}
\definecolor{jttB}{HTML}{1BAF7A}
\definecolor{figsupport}{RGB}{176,240,229}
\definecolor{fignonsupport}{RGB}{249,231,225}
\definecolor{figpatch}{RGB}{217,217,217}
\definecolor{trackred}{HTML}{D62728}
\definecolor{trackdarkgreen}{HTML}{2CA02C}
\definecolor{trackpeach}{HTML}{FFBB78}
\newcommand{\hlbox}[2]{{\setlength{\fboxsep}{1pt}\colorbox{#1}{\vphantom{Tg}#2}}}
\newcommand{\cyanbox}{\hlbox{figsupport}{cyan}}
\newcommand{\pinkbox}{\hlbox{fignonsupport}{pink}}
\newcommand{\graybox}{\hlbox{figpatch}{gray}}
\newcommand{\nickname}{\textsc{JtT}\xspace}
\newcommand{\bv}{\boldsymbol{v}}

\newcommand{\Image}{I}
\newcommand{\sequence}{\boldsymbol{x}}
\newcommand{\query}{q}
\newcommand{\queries}{\mathcal{Q}}
\newcommand{\point}{x}
\newcommand{\velocity}{\Delta}

\newcommand{\eg}{e.g.,\ }

\makeatletter
\renewcommand{\paragraph}{%
  \@startsection{paragraph}{4}%
  {\z@}{0em}{-0.5em}%
  {\normalfont\normalsize\bfseries}%
}
\makeatother

\newcommand{\TA}{Track2Act\xspace}
\newcommand{\WHN}{WHN\xspace}
\newcommand{\TL}{TrajLoom\xspace}
\newcommand{\MF}{MotionForecasting\xspace}
\newcommand{\CoT}{CoTracker3\xspace}
\newcommand{\DRT}{D4RT\xspace}

\newcommand{\TraMoE}{Tra-MoE\xspace}
\newcommand{\GPTF}{GPTF\xspace}

\usepackage{amsmath,amsfonts,bm}

\def\eqref#1{equation~\ref{#1}}

\def\1{\bm{1}}

\DeclareMathAlphabet{\mathsfit}{\encodingdefault}{\sfdefault}{m}{sl}
\SetMathAlphabet{\mathsfit}{bold}{\encodingdefault}{\sfdefault}{bx}{n}

\iclrfinalcopy 

\title{\nickname: Rethinking Point Trajectory Diffusion\\ for Diverse and Rigid Motion from an Image}

\usepackage{marvosym} 

\author{Gabrijel Boduljak\textsuperscript{\Letter} \quad Yushi Lan \quad Christian Rupprecht \quad Andrea Vedaldi \\
\textmd{Visual Geometry Group, University of Oxford}}

\begin{document}
\maketitle
{\let\thefootnote\relax\footnotetext{\textsuperscript{\Letter}Correspondence to: \texttt{gabrijel@robots.ox.ac.uk}}}
\lhead{Preprint. Under review.} 

\begin{figure}[htb]
\centering
\includegraphics[width=0.88\textwidth, keepaspectratio]{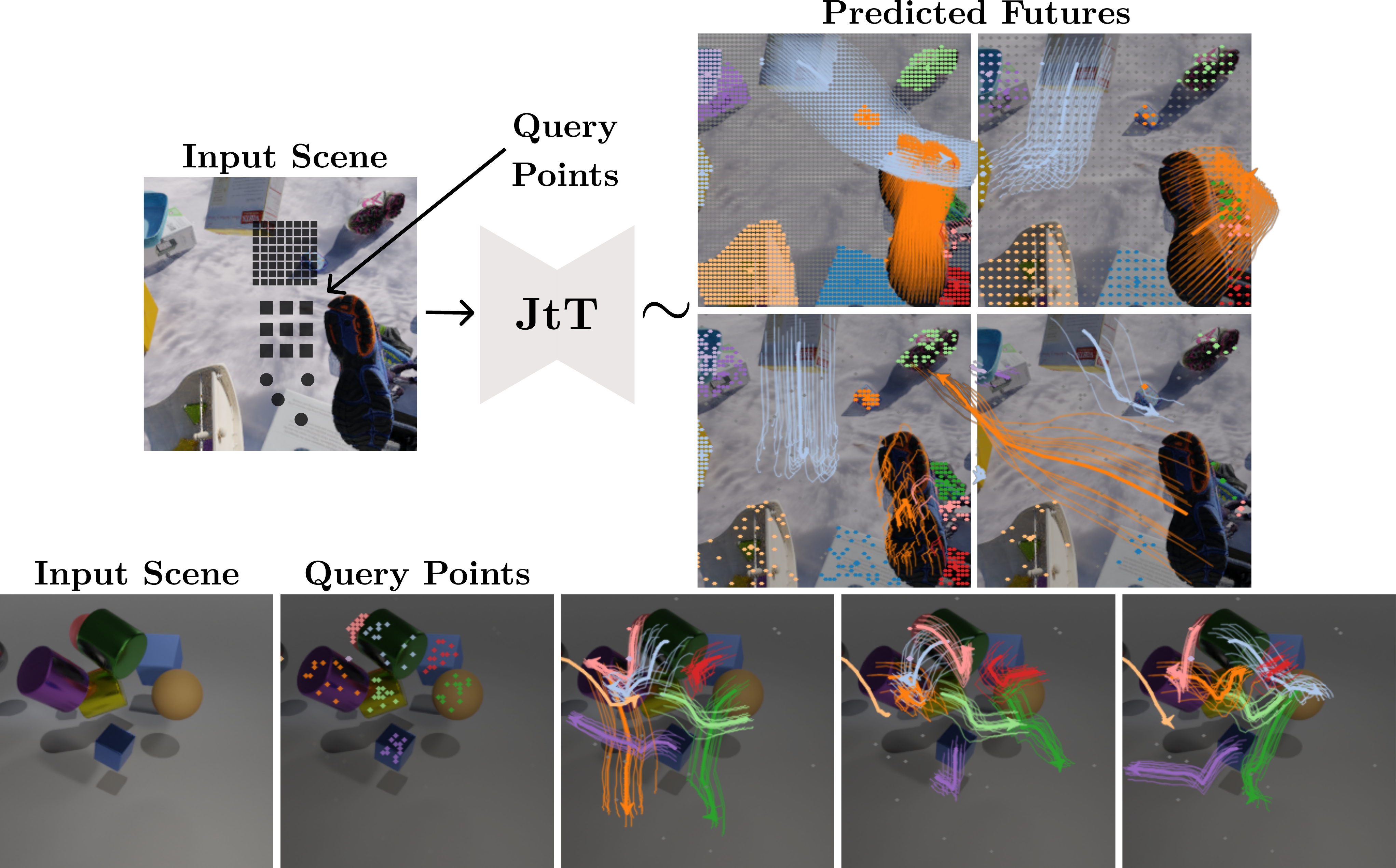}
\caption{Our method, \nickname, infers diverse, physically plausible motion for an arbitrary number of query points, supporting flexible query patterns across diverse camera viewpoints and scene settings.}%
\label{fig:teaser}
\end{figure}

\begin{abstract}
We study forecasting rigid-body motion from a single image: generating diverse and physically plausible future trajectories for arbitrary points in a scene, without building rigidity into the architecture or relying on geometric inputs such as depth.
Like video generators, recent trajectory generators fail at this task, even in simple synthetic Kubric scenes, despite scaling up model size and training on real data.
We show how to solve this with the right architecture and parameterization.
Our model, \nickname, is a simple diffusion transformer that operates directly in coordinate space and decodes arbitrary and dense query patterns, without the trajectory VAEs or fixed grids assumed in prior work.
Through a systematic study of the design space of point trajectory diffusion, we find that diffusing per-frame velocities is the only parameterization that captures rigidity, that it must be paired with the right timestep schedule, and that adding full-resolution RGB at the query points resolves the boundary failures of prior methods.
However, evaluating such 2D models rigorously is itself non-trivial.
To our knowledge, we are the first to observe that physical plausibility and diversity do not improve together during training: larger models, as used in prior work, memorize the single ground-truth future per training image, becoming more rigid while covering fewer possible futures.
Yet prior work has paid little attention to this, so we show how model rankings depend on model size, metric, and stopping time, and adopt a stricter, checkpoint-averaged evaluation protocol.
Under this protocol, \nickname improves over prior methods on both distributional and rigidity metrics and approaches ground-truth rigidity, with only about 50M parameters.
\end{abstract}


\section{Introduction}%
\label{sec:intro}

We study the problem of forecasting the motion of rigid objects from a single image.
These predictors are increasingly used in robot manipulation to generate target motions for policies to execute, represented either as videos~\citep{ko2023learningactactionlessvideos,bharadhwaj2024gen2act,patel2025roboticmanipulationimitatinggenerated} or as point trajectories~\citep{wen2024anypointtrajectorymodelingpolicy,bharadhwaj24track2act:,xu2024im2flow2act,yuan2024generalflow}.
Motion forecasting is also used in model-based control~\citep{yang25tra-moe:,mazzaglia24genrl:} and controllable video generation~\citep{polyak24movie,wan2025,brooks24video}.


Since an image does not fully determine the state of the scene, many futures are possible, and a useful predictor must capture their distribution rather than a single realization.
The goal is to make such realizations physically plausible.
Unfortunately, even the best generative forecasting models struggle with physical plausibility.
Video generators often violate object permanence and basic physical laws~\citep{kang25how-far-is-video,motamed25do-generative,bansal25videophy:,bansal26videophy-2:};
in particular, they may fail to predict rigid motion even when fine-tuned for forecasting in simple synthetic scenes~\citep{boduljak2026whn}.
Some authors simply assume that objects are rigid and only predict rigid motion~\citep{ko2023learningactactionlessvideos,bharadhwaj24track2act:,patel2025roboticmanipulationimitatinggenerated}, but this is not general.

Using 2D point trajectories as the representation for motion forecasting has several advantages:
it is general, it guarantees object permanence and temporal coherence by construction, it avoids the need to recover 2D motion from generated pixels~\citep{ko2023learningactactionlessvideos,patel2025roboticmanipulationimitatinggenerated}, and it yields more rigid motion than videos under an identical architecture~\citep{boduljak2026whn}.
Furthermore, as for video models, large quantities of real-world training data can be obtained by using reliable 2D point trackers~\citep{doersch2023tapirtrackingpointperframe,doersch24bootstap:,karaev25cotracker3} for pseudo-labeling videos~\citep{bharadhwaj24track2act:,zhang2026trajloomdensefuturetrajectory,thakkar2026forecasting,lu2026generative}.

However, point trajectories alone do not guarantee plausible motion: even on the synthetic Kubric-A scenes, which contain only a few rigid objects, no prior trajectory generator achieves exact rigidity (\cref{tab:overall}).
Instead, they trade off plausibility and diversity:
\MF{}~\citep{thakkar2026forecasting} is the most rigid but has the worst distributional fidelity, whereas \WHN{}~\citep{boduljak2026whn} is the opposite.
They also often miss small objects and object parts and generate jittery backgrounds (\cref{fig:comparison}).

In this paper, we study in depth how to learn to predict diverse and plausible 2D point trajectories from single images.
We deliberately avoid injecting hard inductive biases.
We also avoid using additional observations of the scene, such as depth.
We consider this to be a particularly interesting case for two reasons.
First, it is a hard setting for the model: without additional observations, the distribution of possible futures is highly multimodal and difficult to capture even using diffusion~\citep{lavoie2025compositional}.
Second, 2D pseudo-labels can be obtained with reasonable accuracy by modern trackers~\citep{doersch2023tapirtrackingpointperframe,doersch24bootstap:,karaev25cotracker3}, but 3D pseudo-labels remain much less accurate~\citep{koppula24tapvid-3d:,xiao24spatialtracker:,xiao25spatialtrackerv2:} and harder to get, which makes our case of practical interest.


We are not the first to consider this an interesting setting.
However, we have identified two gaps in current best practices that we believe are limiting progress in this area.
The first is a modeling gap and the second is an evaluation gap.
Our contribution is to characterize and address them.

On the \textbf{modeling gap}, existing trajectory generators (\cref{tab:trajectory_eval}) compromise on at least one of three quality axes:
(i) \emph{physical plausibility}, respecting physical invariants such as the rigidity of solid objects;
(ii) \emph{query flexibility}, querying any number of points at arbitrary 2D locations at inference; and
(iii) \emph{query density}, predicting motion everywhere in the image rather than at a sparse, fixed set of points.
%
%
The default approach to improving quality is to scale the model and training data, or to introduce latent RGB and video representations, or custom VAE losses and rollout fine-tuning~\citep{boduljak2026whn,zhang2026trajloomdensefuturetrajectory}.
Here we take a step back and ask whether a simple but well-designed diffusion model that operates directly in trajectory (coordinate) space can do better.
We show it can, given four design choices that prior works treat inconsistently:
(1) what to diffuse: absolute coordinates, offsets relative to the query, or velocities;
(2) how to inject visual information: sampled image features or full-resolution RGB;
(3) which timestep schedule to use for flow matching; and
(4) which inductive biases help, in particular implicit (\eg reconstruction model features) or explicit (\eg world coordinates) 3D geometry.
Most importantly, diffusing per-frame velocities is the only parameterization that captures rigidity, and adding full-resolution RGB at the query points resolves the boundary failures of prior methods, which use low-resolution image features (\cref{fig:comparison}).

Building on these findings, we introduce the \emph{Just trajectory Transformer} (\nickname; \cref{fig:method}), which predicts the per-frame velocities of arbitrary query points in an input frame.
A small \emph{support set} of query tokens does the heavy lifting, attending to each other and to multi-layer image features, while all other \emph{non-support} queries decode their trajectories cheaply by cross-attending the support set.
Unlike Perceiver IO~\citep{jaegle2021perceiver,jaegle22perceiver}, \DRT{}~\citep{zhang26efficiently}, and \CoT{}~\citep{karaev25cotracker3}, which use latent or support tokens for \textit{tracking}, we use them in a generative model and randomly re-partition the queries at every training step, so that a single model, trained with 1 to 1024 query points, decodes arbitrary and dense query patterns without retraining (\cref{fig:teaser}) and lets users trade compute for accuracy.
\nickname requires no VAE, auxiliary losses, rollout fine-tuning, or pre-trained video latents; with only about 50M parameters, it outperforms prior methods on FVMD and rigidity.

On the \textbf{evaluation gap},
we note that assessment in prior works is often unreliable, partly due to reporting a single checkpoint when model ranking can change substantially from one to the next, and partly due to only reporting results on real videos, which, each providing a single realization, make it difficult to measure distributional coverage.
In general, rigorous evaluation is non-trivial.
For instance, rigidity cannot be measured directly from 2D point trajectories and must be assessed via proxy metrics.
Furthermore, there is a clear trade-off between physical plausibility and distributional coverage.

By studying the training dynamics of trajectory generators in a controlled setting, to our knowledge for the first time, we find that larger models converge faster but then memorize the single ground-truth future per training image, becoming more rigid while covering fewer possible futures (\cref{fig:dynamics}), so that which model appears best depends on model size, metric, and stopping time.
We therefore evaluate with a checkpoint-averaged protocol on the Kubric-A benchmark of~\citet{boduljak2026whn}, whose multiple ground-truth futures per image enable rigorous evaluation of stochastic predictors; on real videos, a model that memorizes the single observed future (or predicts the implausible average) would even appear best.


We summarize our contributions as follows:
(1) We introduce \nickname, a transformer architecture and diffusion parameterization that learns the 2D motion of rigid bodies with neither rigidity built into the architecture nor extra geometric inputs, while decoding arbitrary and dense query patterns.
It improves over prior methods on distributional and rigidity metrics, approaching ground-truth rigidity, while using a smaller model and simpler design than many prior works.
(2) We systematically study the design space of point trajectory diffusion and show that diffusing per-frame velocities is the only parameterization that captures rigidity, that it depends on the right timestep schedule, and that full-resolution RGB at the query points resolves boundary failures; these choices should take precedence over more complex model architectures, such as introducing trajectory VAEs.
(3) To our knowledge, we are the first to show that physical plausibility and diversity do not improve together during the training of point trajectory generators: larger models converge faster but then memorize the single ground-truth future per training image, improving rigidity while degrading distributional fidelity, so that the ranking of models depends on model size and on the evaluated checkpoint.
We therefore adopt a checkpoint-averaged protocol for rigorous evaluation.
(4) We show that explicit oracle geometry, such as inverse depth at query points or generation in world coordinates, improves performance, but, in terms of rigidity, our 2D-only model closes most of the gap between prior 2D methods and the 3D oracle, verifying our design; integrating 3D reconstruction models is promising future work.

\section{Related work}%
\label{sec:related}

\paragraph{Visual trajectory forecasting.}

Visual trajectory forecasting has progressed quickly, but existing methods have limitations that our work addresses.
\textbf{\TA{}}~\citep{bharadhwaj24track2act:} is a 600M-parameter diffusion model that predicts absolute coordinates for arbitrary query points, but it conditions on a single ResNet feature vector, which limits spatial resolution and causes it to hallucinate objects and miss small ones.
\textbf{\WHN{}}~\citep{boduljak2026whn} runs latent rectified flow over a fixed $32{\times}32$ grid via a VAE with patch-level DINO conditioning, improving diversity and showing the importance of global reasoning over large point sets.
However, the fixed grid prevents arbitrary query patterns, the VAE increases training complexity, and jitter persists in generated samples, even on Kubric-A, with degraded accuracy near boundaries.
\textbf{\TL{}}~\citep{zhang2026trajloomdensefuturetrajectory} extends \WHN{}'s latent flow matching to video-conditioned generation at a larger scale.
It diffuses VAE latents on a fixed $15{\times}26$ grid conditioned on Wan2.1~\citep{wan2025} RGB-video latents, and introduces custom VAE losses along with rollout fine-tuning.
\textbf{\MF{}}~\citep{thakkar2026forecasting}
adapts \TA{} to wildlife footage by removing the goal image and replacing the ResNet with DINOv3~\citep{simeoni25dinov3}.
The method samples low-resolution image features, resulting in failure modes similar to those of \TA{}.
\textbf{\GPTF{}}~\citep{lu2026generative} unifies point tracking and forecasting in one generative model, diffusing absolute coordinates given DINOv3 patch features.

\newcommand{\cmark}{\ding{51}}
\newcommand{\xmark}{\ding{55}}
\begin{table}[t]
\vspace{-3mm}
\centering
\small
\caption{\textbf{Design space of image-conditioned trajectory generators.}
\emph{Queries}: where trajectories can be requested at inference;
\emph{Dense}: decodes grids denser than in training;
\emph{Space}: where diffusion is performed;
\emph{Target}: the diffused trajectory representation; $^*$: estimated (\cref{app:gptf-params}).}%
\label{tab:trajectory_eval}
\setlength{\tabcolsep}{3.4pt}
\begin{tabular}{@{}l cc ll l r@{}}
\toprule
& \multicolumn{2}{c}{\textbf{Flexibility}} & \multicolumn{2}{c}{\textbf{Representation}} & & \\
\cmidrule(lr){2-3} \cmidrule(lr){4-5}
\textbf{Method} & \textbf{Queries} & \textbf{Dense} & \textbf{Space} & \textbf{Target} & \textbf{Image conditioning} & \textbf{Parameters} \\
\midrule
\TA{} & arbitrary           & \xmark & coords. & absolute $\point_t$              & global vector& 600M \\
\WHN{} & fixed $32{\times}32$ & \xmark & VAE latent  & absolute $\point_t$              & DINOv2 patches& 100/250M \\
\TL{} & fixed $15{\times}26$ & \xmark & VAE latent  & offset $\point_t - \point_c$ & Wan2.1 latents& 250M \\
\MF{} & arbitrary           & \xmark & coords. & velocity $\velocity_t$           & DINOv3 at query & 100M \\
\GPTF{} & arbitrary & \xmark & coords. & absolute $\point_t$ & DINOv3 patches & 30/230M$^*$ \\
\midrule
\textbf{\nickname (ours)} & \textbf{arbitrary} & \cmark & \textbf{coords.} & \textbf{velocity} $\bm{\velocity_t}$ & \textbf{RGB + DINO patches} & \textbf{50/220M} \\
\bottomrule
\end{tabular}
\vspace{-7.6mm}
\end{table}

\nickname differs from these methods in three ways.
First, we operate directly in coordinate space without any VAE\@.
Second, we condition on full-resolution RGB and multi-layer DINO features rather than low-resolution global features or RGB-video VAE latents.
Third, we decouple the query distribution from the architecture via the support/non-support partition, so the same model can predict with arbitrary configurations of query points, including dense query patterns (\cref{fig:teaser}).

\paragraph{Architectures supporting arbitrary configurations of query points.}

An important property of our design is support for arbitrary query configurations.
Our design is inspired by Perceiver~\citep{jaegle2021perceiver} and Perceiver IO~\citep{jaegle22perceiver}, which introduce the idea of using a latent representation of fixed size paired with an arbitrary collection of output queries to read it out.
This decouples the size of model inputs, internal state, and outputs; \DRT{}~\citep{zhang26efficiently} applies this idea to dynamic 4D reconstruction with randomly resampled read-write tokens.
Closer to our domain, in the context of 2D tracking, \CoT{}~\citep{karaev25cotracker3} (and its predecessor~\citep{karaev24cotracker}) use a small set of \emph{support} tracks that interact through self-attention and a large set of \emph{non-support} tracks that read from them.
Our support/non-support decomposition (\cref{sec:arch}) uses a similar idea for \emph{generating} instead of estimating tracks.
Unlike in prior work, the partition is randomized at every training step to keep the model general.
Thus, a single model trained with $1$ to $1{,}024$ points per scene generalizes to dense query patterns at inference time.
\section{Method}%
\label{sec:method}

\begin{figure}[ht]
\centering
\includegraphics[width=\textwidth, keepaspectratio]{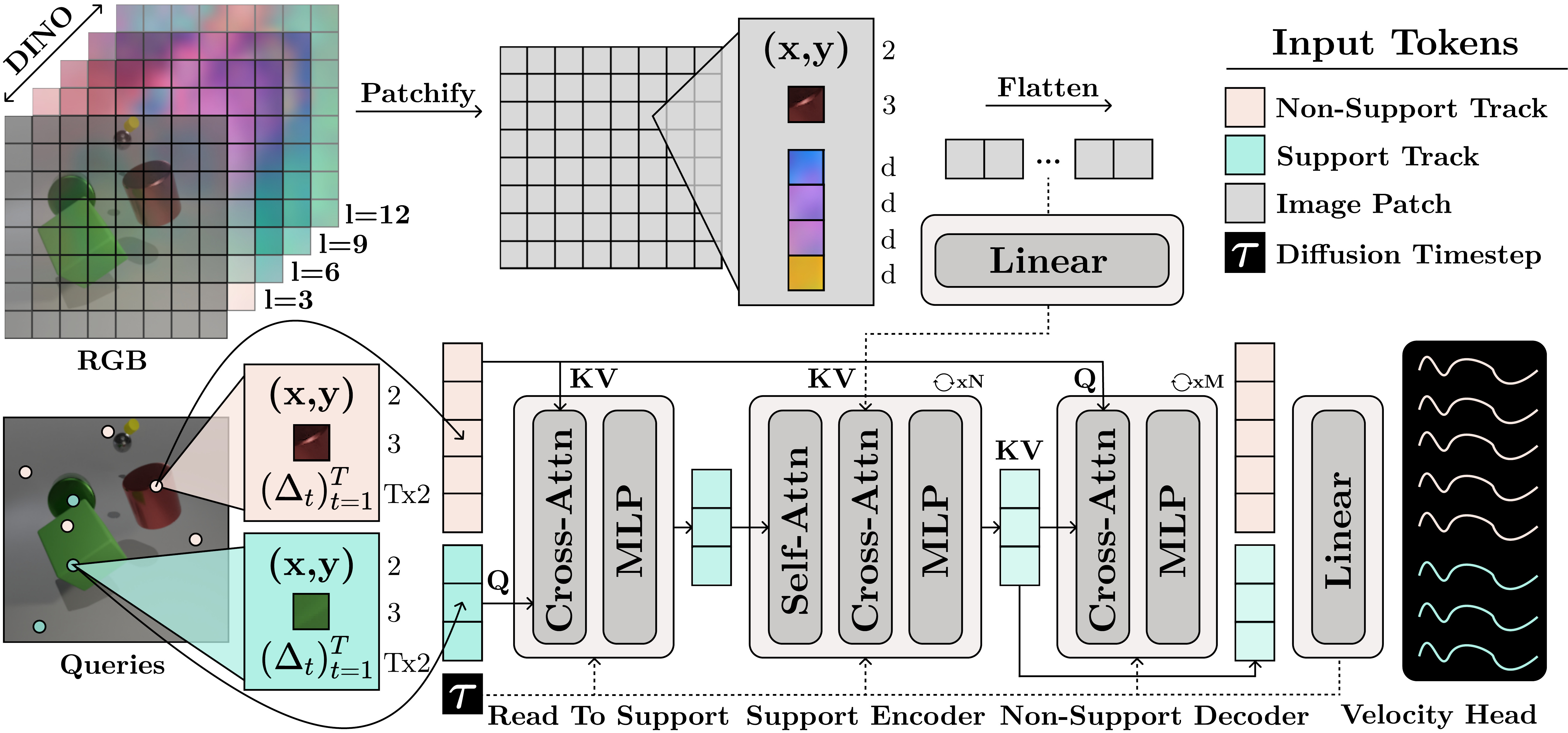}
\caption{\textbf{Overview of \nickname.}
Given an input image and query points, image patch tokens (\graybox) are formed by concatenating patch position $(x,y)$, RGB content, and multi-layer DINOv2 features ($l \in \{3,6,9,12\}$), while query points are tokenized analogously with position, sampled RGB, and noised per-frame velocities $(\Delta_t)_{t=1}^{T}$, and then split into a support set (\cyanbox) and a non-support set (\pinkbox).
A \emph{Read-To-Support} cross-attention block initializes support tokens from non-support tokens.
The \emph{Support Encoder} ($\times N$) refines the support set via interleaved self-attention and cross-attention to image patches.
The \emph{Non-Support Decoder} ($\times M$) reads out from the encoded support set in parallel; non-support tokens never attend to each other or back to the image, making dense readout linear in $M$ and reusable for any query pattern at inference.
A linear \emph{Velocity Head} predicts the rectified-flow velocities, which are accumulated to recover trajectories.
All blocks are AdaLN-modulated by the diffusion timestep $\tau$.}%
\label{fig:method}
\vspace{-4mm}
\end{figure}


We study the problem of visual trajectory forecasting.
Formally, a trajectory is a sequence
$
\sequence = (\point_0, \dots, \point_T) \in \mathbb{R}^{(T+1) \times 2}
$
of 2D points
$
\point_t = (u_t, v_t) \in \mathbb{R}^2
$
that starts at a query location
$
\query \equiv (u_0, v_0)
$
(i.e., $\point_0 = \query$).
The model is given as input an image
$
\Image \in \mathbb{R}^{H \times W \times 3}
$
as well as a collection of $N$ query points
$
\queries = \{\query_n\}_{n=1}^N
$.
The goal of the model is to sample corresponding trajectories from the conditional distribution
$
p(\{\sequence_n\}_{n=1}^N \mid \Image, \queries)
$.
Our problem formulation is more flexible than some prior works; in particular, compared to~\citet{boduljak2026whn}, the set of query points $\queries$ is supplied at inference rather than being fixed.
On the other hand, unlike~\citet{bharadhwaj24track2act:}, the model is only given the image $\Image$ as evidence; it is not given a goal frame, segmentation, or correspondences.

We introduce a new model, \nickname, to solve this problem.
Our design, illustrated in \cref{fig:method}, is
(i) \emph{generative}, sampling multiple distinct futures per scene;
(ii) \emph{flexible}, accepting an arbitrary number of query points at arbitrary 2D locations; and
(iii) \emph{dense}, scaling to thousands of points per scene on a single GPU\@.
We take a deliberately minimal approach and train a single diffusion transformer that operates directly in coordinate space.
In particular, we use no VAEs, auxiliary losses, or pre-trained video latents.
While the design is simple, we find numerous choices critical to achieving good performance.
We discuss these next.




\subsection{Trajectory parameterization}%
\label{sec:param}

We parameterize each trajectory as the sequence of \emph{instantaneous velocities}
$
\velocity_t \equiv \point_t - \point_{t-1} \in \mathbb{R}^2$, $t = 1, \ldots, T
$
and target those for generation using denoising diffusion.
We isotropically normalize the velocities across the dataset so the diffusion target has roughly unit variance.
The full trajectory for query $\query$ is recovered at inference as
$
\point_t  = \query + \sum_{i=1}^{t} \velocity_i
$.
Compared with \emph{absolute coordinates} $x_t$ (\TA{}, \WHN{}), which couple the diffusion target to the query location, and \emph{grid-cell offsets} $x_t - x_c$ (TrajLoom), which assume a fixed inference grid, per-frame velocities are translation-equivariant in the query, place the target on a near-zero, dataset-normalized scale, and decouple the trajectory from the absolute pixel grid.
Empirically, we found that this helps the model learn the motion of rigid objects under arbitrary query patterns.

\subsection{Image conditioning}%
\label{sec:cond}

We patchify the image $\Image$ into a regular $P \times P$ grid and form one token per patch by concatenating
(i) the 2D pixel center $(u, v) \in \mathbb{R}^2$,
(ii) the RGB patch itself (stacked to a $3P^2$ vector), and
(iii) DINOv2~\citep{oquab2023dinov2} features extracted from \emph{multiple} transformer layers $l \in \{3, 6, 9, 12\}$, each contributing a $d$-dimensional descriptor, analogous to the multi-layer aggregation in DPT~\citep{ranftl21vision}.
A single linear layer maps these vectors to a fixed hidden dimension, yielding $P^2$ image tokens that the trajectory tokens cross-attend to.
We found that including full-resolution RGB patches instead of a single global vector (\TA{}) or only a low-resolution feature grid (\MF{}) works better, particularly at image boundaries, and reduces hallucinated and missed objects that sometimes affect prior work.
We also observed that multi-layer DINO features perform better than relying solely on the final layer.

\subsection{Architecture}%
\label{sec:arch}

The \nickname model (\cref{fig:method}) is a diffusion transformer that takes as input the trajectory velocity $\Delta_\tau$ at a noise level $\tau$, the image $\Image$, and the query set $\queries$, and outputs a denoising signal $\hat{\bv}_\theta(\velocity_\tau, \Image, \queries, \tau)$ (see \cref{sec:train}).
Hence, the transformer must process the velocities $\velocity$ encoded by the query tokens as well as the input image $\Image$ and noise level $\tau$.

The bulk of the computation is taken up by the query tokens, for which we use a Perceiver-IO-style~\citep{jaegle2021perceiver,jaegle22perceiver} encoder-decoder.
The idea is to split the $N$ query points $\queries$ into a \emph{support} set of $S$ tokens (\cyanbox{} in the figure) and a \emph{non-support} set of $M = N - S$ tokens (\pinkbox).
Support tokens attend to the image features and reason globally; non-support tokens are decoded cheaply by reading information off the support set in parallel.
By comparison, self-attention across all query tokens would scale quadratically in their number $N$, which would make dense prediction (e.g., using a $256 \times 256$ grid of queries) all but intractable.
CoTracker3~\citep{karaev25cotracker3} and D4RT~\citep{zhang26efficiently} sidestep this problem by using a latent representation of fixed dimensionality, which acts as a bottleneck.
We retain that efficiency, but make the support set fully flexible.
Users can specify any set of queries $\queries$ and still control the compute budget by choosing the size $S\ll N$ of the support set, without having to retrain the model.
The diffusion timestep $\tau$ modulates layer norms via AdaLN~\citep{peebles2023scalablediffusionmodelstransformers} in every block.

\subsection{Training and sampling}%
\label{sec:train}

We train \nickname with a rectified-flow (stochastic-interpolant) objective~\citep{lipman22flow,liu23flow,albergo2023stochastic}.
With ground-truth velocities $\velocity_1 \in \mathbb{R}^{T \times 2}$ (isotropically normalized as noted above) and $\velocity_0 \sim \mathcal{N}(0, I)$, the linear diffusion path is $\velocity_\tau = (1 - \tau)\velocity_0 + \tau \velocity_1$ with constant diffusion velocity $\bv = \velocity_1 - \velocity_0$ (not to be confused with the trajectory velocity).
The model $\hat{\bv}_\theta(\velocity_\tau, \Image, \queries, \tau)$ is trained by minimizing
\[
\mathcal{L}(\theta) = \mathbb{E}\left[ \big \| \hat{\bv}_\theta(\velocity_\tau, \Image, \queries, \tau) - (\velocity_1 - \velocity_0) \big \|_2^2 \right],
\]
with $\tau$ sampled from a logit-normal distribution~\citep{esser2024scalingrectifiedflowtransformers} rather than uniformly: a uniform schedule overfits the single observed future per initial condition, while logit-normal up-weights mid-noise levels and yields sharper, more diverse samples (\cref{tab:diffusion-results}).

For each video, we take the initial frame, sample $N \sim \mathcal{U}[1, 1024]$ pixel queries, and use the next $T = 24$ frames as ground truth.
For each $N$, tokens are randomly partitioned into a support set ($S \le S_{\max}$) and a non-support set ($N - S$).
This randomization is a strong inductive prior: trajectories must be consistent under arbitrary subsampling and orderings, and any token may be promoted to or demoted from the support role.
We use no data augmentation, rollout fine-tuning, or auxiliary objectives.
At inference, we draw $\velocity_0 \sim \mathcal{N}(0, I)$ and solve the SDE with the learned drift via 1000 Euler--Maruyama steps, then accumulate the predicted velocities.
We use this sampler because it gives the best rigidity (LRTL); a 10-step Euler ODE sampler achieves similar rigidity and even better scores on the other metrics (\cref{tab:sampling}).
Details are in \cref{app:impl}.

\section{Experiments}%
\label{sec:experiments}

We begin by studying the interplay between model size, sample diversity, and physical plausibility during training, which determines how models should be selected and compared.
We then show that \nickname outperforms prior work, explain step by step which design choices lead to this performance, and finally compare it with oracles given privileged 3D geometry.

\paragraph{Data.}
We run most experiments on {Kubric-A}, strictly following the protocol of \WHN{}~\citep{boduljak2026whn}: sequences of $T = 24$ frames rendered with the Kubric simulator~\citep{greff2021kubric}, with multiple ground-truth futures per initial condition obtained by re-simulating the physics under different initial velocities.
As established in \WHN{}, this setup is necessary for rigorous evaluation of a stochastic motion predictor: real videos supply only one future per scene and no instance-level rigidity ground truth, ruling out both distributional metrics and physically grounded ones such as LRTL.
We also use {Kubric-D}, with richer 3D scene geometry, to assess explicit geometric inputs.
We verify our findings on diversity and rigidity qualitatively on real videos from Cityscapes~\citep{cordts16the-cityscapes} and Physics 101~\citep{physics101} (\cref{app:qualitative-physics101,app:qualitative-cityscapes}; videos in the supplement).

\paragraph{Metrics.}
We follow the evaluation protocol of~\citet{boduljak2026whn}; lower is better for all metrics.
Since motion prediction from a single image is multimodal, we report the \emph{Best-of-$K$} MSE between $K$ generated and $K$ simulated futures per image.
We measure distributional fidelity with \emph{FVMD}~\citep{liu2024fr}, a Fréchet distance between trajectory features, computed over the whole dataset (D) and per image (S).
Directly evaluating rigidity is impossible for 2D outputs, since the image-plane projection of a rigid motion is not itself rigid; we therefore measure physical plausibility with the proxy metric \emph{LRTL}~\citep{karazija24learning,boduljak2026whn}, the rank-5 reconstruction error of trajectories grouped by object using simulated masks, which is low for rigid motion.
FVMD and LRTL are complementary: FVMD can miss local inconsistencies such as shuffled velocities, while LRTL is minimized by static trajectories.

\begin{figure}[t]
\centering
\begin{tikzpicture}
\begin{groupplot}[
group style={group size=2 by 2, horizontal sep=1.35cm, vertical sep=0.55cm},
width=0.5\textwidth, height=0.3\textwidth,
xmin=35, xmax=135, xtick={40,70,100,130}, minor xtick={50,60,80,90,110,120}, xticklabel={\pgfmathprintnumber{\tick}K},
tick label style={font=\footnotesize}, title style={font=\small, yshift=-6pt},
scaled y ticks=false, yticklabel style={/pgf/number format/.cd, fixed, 1000 sep={\mathord{,}}},
axis line style={black!45}, tick style={black!45}, grid=major, grid style={black!8, line width=0.3pt},
legend style={font=\small, draw=none, fill=none, legend columns=-1, column sep=10pt},
]
\nextgroupplot[title={FVMD (D) $\downarrow$}, xticklabels={}, legend style={at={(1.1,1.16)}, anchor=south}]
\addplot[jttXS, line width=0.4pt, opacity=0.45, mark=*, mark size=1.3pt, forget plot] coordinates {(40,18319.062) (50,14164.582) (60,13091.881) (70,13019.896) (80,12975.826) (90,13004.096) (100,12643.575) (110,12389.515) (120,12244.834) (130,11835.232)};
\addplot[jttXS, line width=1.3pt, mark=none, legend image post style={mark=*, mark size=1.8pt}] coordinates {(40,16241.822) (50,15191.842) (60,13425.453) (70,13029.201) (80,12999.939) (90,12874.499) (100,12679.062) (110,12425.975) (120,12156.527) (130,12040.033)};
\addlegendentry{\nickname-XS}
\addplot[jttS, line width=0.4pt, opacity=0.45, mark=square*, mark size=1.3pt, forget plot] coordinates {(40,14093.199) (50,14286.340) (60,13068.819) (70,12237.690) (80,12328.595) (90,12746.655) (100,12349.562) (110,14276.285) (120,13286.094) (130,12817.273)};
\addplot[jttS, line width=1.3pt, mark=none, legend image post style={mark=square*, mark size=1.8pt}] coordinates {(40,14189.770) (50,13816.119) (60,13197.616) (70,12545.035) (80,12437.647) (90,12474.937) (100,13124.167) (110,13303.980) (120,13459.884) (130,13051.683)};
\addlegendentry{\nickname-S}
\addplot[jttB, line width=0.4pt, opacity=0.45, mark=triangle*, mark size=1.3pt, forget plot] coordinates {(40,12017.968) (50,11675.977) (60,11812.407) (70,11957.168) (80,14304.908) (90,14615.274) (100,14880.139) (110,14767.617) (120,15136.429) (130,15708.267)};
\addplot[jttB, line width=1.3pt, mark=none, legend image post style={mark=triangle*, mark size=1.8pt}] coordinates {(40,11846.973) (50,11835.451) (60,11815.184) (70,12691.494) (80,13625.783) (90,14600.107) (100,14754.343) (110,14928.062) (120,15204.104) (130,15422.348)};
\addlegendentry{\nickname-B}
\addlegendimage{black!60, line width=1.0pt}
\addlegendentry{train loss}
\addlegendimage{black!60, line width=1.0pt, dashed}
\addlegendentry{validation loss}
\draw[jttXS, densely dotted, line width=0.8pt] ({axis cs:130,0} |- {rel axis cs:0,0}) -- ({axis cs:130,0} |- {rel axis cs:0,1});
\draw[jttXS, line width=0.8pt] (axis cs:130,11835.232) circle[radius=3pt];
\draw[jttS, densely dotted, line width=0.8pt] ({axis cs:100,0} |- {rel axis cs:0,0}) -- ({axis cs:100,0} |- {rel axis cs:0,1});
\draw[jttS, line width=0.8pt] (axis cs:100,12349.562) circle[radius=3pt];
\draw[jttB, densely dotted, line width=0.8pt] ({axis cs:60,0} |- {rel axis cs:0,0}) -- ({axis cs:60,0} |- {rel axis cs:0,1});
\draw[jttB, line width=0.8pt] (axis cs:60,11812.407) circle[radius=3pt];
\nextgroupplot[title={FVMD (S) $\downarrow$}, xticklabels={}]
\addplot[jttXS, line width=0.4pt, opacity=0.45, mark=*, mark size=1.3pt, forget plot] coordinates {(40,21504.536) (50,18125.482) (60,17539.278) (70,17629.147) (80,17532.914) (90,17595.794) (100,17117.613) (110,16824.290) (120,16598.983) (130,16084.196)};
\addplot[jttXS, line width=1.3pt, mark=none, legend image post style={mark=*, mark size=1.8pt}] coordinates {(40,19815.009) (50,19056.432) (60,17764.636) (70,17567.113) (80,17585.952) (90,17415.440) (100,17179.232) (110,16846.962) (120,16502.490) (130,16341.589)};
\addplot[jttS, line width=0.4pt, opacity=0.45, mark=square*, mark size=1.3pt, forget plot] coordinates {(40,18905.157) (50,18976.823) (60,17512.305) (70,16350.717) (80,16447.447) (90,16941.527) (100,16597.454) (110,18668.485) (120,17726.122) (130,17244.561)};
\addplot[jttS, line width=1.3pt, mark=none, legend image post style={mark=square*, mark size=1.8pt}] coordinates {(40,18940.990) (50,18464.762) (60,17613.282) (70,16770.156) (80,16579.897) (90,16662.143) (100,17402.489) (110,17664.020) (120,17879.722) (130,17485.341)};
\addplot[jttB, line width=0.4pt, opacity=0.45, mark=triangle*, mark size=1.3pt, forget plot] coordinates {(40,16129.730) (50,15606.083) (60,15828.663) (70,16042.628) (80,18627.156) (90,18911.210) (100,19169.120) (110,18905.608) (120,19413.616) (130,20181.348)};
\addplot[jttB, line width=1.3pt, mark=none, legend image post style={mark=triangle*, mark size=1.8pt}] coordinates {(40,15867.907) (50,15854.826) (60,15825.791) (70,16832.815) (80,17860.331) (90,18902.495) (100,18995.313) (110,19162.782) (120,19500.191) (130,19797.482)};
\draw[jttXS, densely dotted, line width=0.8pt] ({axis cs:130,0} |- {rel axis cs:0,0}) -- ({axis cs:130,0} |- {rel axis cs:0,1});
\draw[jttXS, line width=0.8pt] (axis cs:130,16084.196) circle[radius=3pt];
\draw[jttS, densely dotted, line width=0.8pt] ({axis cs:100,0} |- {rel axis cs:0,0}) -- ({axis cs:100,0} |- {rel axis cs:0,1});
\draw[jttS, line width=0.8pt] (axis cs:100,16597.454) circle[radius=3pt];
\draw[jttB, densely dotted, line width=0.8pt] ({axis cs:60,0} |- {rel axis cs:0,0}) -- ({axis cs:60,0} |- {rel axis cs:0,1});
\draw[jttB, line width=0.8pt] (axis cs:60,15828.663) circle[radius=3pt];
\nextgroupplot[title={LRTL (ALL) $\downarrow$}, xlabel={Training step}, xlabel style={font=\footnotesize, yshift=2pt}]
\addplot[jttXS, line width=0.4pt, opacity=0.45, mark=*, mark size=1.3pt, forget plot] coordinates {(40,7.384) (50,8.706) (60,10.854) (70,11.605) (80,11.173) (90,11.202) (100,10.350) (110,9.623) (120,9.112) (130,8.105)};
\addplot[jttXS, line width=1.3pt, mark=none, legend image post style={mark=*, mark size=1.8pt}] coordinates {(40,8.045) (50,8.982) (60,10.388) (70,11.211) (80,11.327) (90,10.908) (100,10.392) (110,9.695) (120,8.947) (130,8.609)};
\addplot[jttS, line width=0.4pt, opacity=0.45, mark=square*, mark size=1.3pt, forget plot] coordinates {(40,13.154) (50,11.378) (60,9.273) (70,7.536) (80,7.512) (90,7.231) (100,6.909) (110,7.063) (120,6.726) (130,6.615)};
\addplot[jttS, line width=1.3pt, mark=none, legend image post style={mark=square*, mark size=1.8pt}] coordinates {(40,12.266) (50,11.268) (60,9.396) (70,8.107) (80,7.426) (90,7.217) (100,7.068) (110,6.899) (120,6.801) (130,6.671)};
\addplot[jttB, line width=0.4pt, opacity=0.45, mark=triangle*, mark size=1.3pt, forget plot] coordinates {(40,9.329) (50,7.691) (60,7.273) (70,6.770) (80,6.827) (90,6.366) (100,6.129) (110,5.578) (120,5.449) (130,5.313)};
\addplot[jttB, line width=1.3pt, mark=none, legend image post style={mark=triangle*, mark size=1.8pt}] coordinates {(40,8.510) (50,8.098) (60,7.245) (70,6.957) (80,6.654) (90,6.441) (100,6.024) (110,5.719) (120,5.447) (130,5.381)};
\nextgroupplot[title={Flow-matching loss $\downarrow$}, xlabel={Training step}, xlabel style={font=\footnotesize, yshift=2pt}, yticklabel style={/pgf/number format/.cd, fixed, precision=2}, ytick={0.03,0.06,0.09,0.12}]
\addplot[jttXS, line width=1.0pt, mark=*, mark size=1.3pt, forget plot] coordinates {(40,0.1267) (50,0.0957) (60,0.0842) (70,0.0780) (80,0.0734) (90,0.0698) (100,0.0669) (110,0.0642) (120,0.0618) (130,0.0597)};
\addplot[jttXS, line width=1.0pt, dashed, mark=*, mark size=1.3pt, mark options={solid}, forget plot] coordinates {(40,0.1302) (50,0.0984) (60,0.0865) (70,0.0802) (80,0.0756) (90,0.0721) (100,0.0692) (110,0.0666) (120,0.0643) (130,0.0623)};
\addplot[jttS, line width=1.0pt, mark=square*, mark size=1.3pt, forget plot] coordinates {(40,0.1242) (50,0.0968) (60,0.0672) (70,0.0544) (80,0.0486) (90,0.0456) (100,0.0430) (110,0.0415) (120,0.0392) (130,0.0376)};
\addplot[jttS, line width=1.0pt, dashed, mark=square*, mark size=1.3pt, mark options={solid}, forget plot] coordinates {(40,0.1276) (50,0.1004) (60,0.0710) (70,0.0590) (80,0.0540) (90,0.0522) (100,0.0512) (110,0.0518) (120,0.0520) (130,0.0534)};
\addplot[jttB, line width=1.0pt, mark=triangle*, mark size=1.3pt, forget plot] coordinates {(40,0.0794) (50,0.0535) (60,0.0432) (70,0.0377) (80,0.0343) (90,0.0307) (100,0.0281) (110,0.0251) (120,0.0225) (130,0.0204)};
\addplot[jttB, line width=1.0pt, dashed, mark=triangle*, mark size=1.3pt, mark options={solid}, forget plot] coordinates {(40,0.0849) (50,0.0616) (60,0.0556) (70,0.0565) (80,0.0610) (90,0.0674) (100,0.0757) (110,0.0869) (120,0.0977) (130,0.1104)};
\draw[jttXS, densely dotted, line width=0.8pt] ({axis cs:130,0} |- {rel axis cs:0,0}) -- ({axis cs:130,0} |- {rel axis cs:0,1});
\draw[jttXS, line width=0.8pt] (axis cs:130,0.0623) circle[radius=3pt];
\draw[jttS, densely dotted, line width=0.8pt] ({axis cs:100,0} |- {rel axis cs:0,0}) -- ({axis cs:100,0} |- {rel axis cs:0,1});
\draw[jttS, line width=0.8pt] (axis cs:100,0.0512) circle[radius=3pt];
\draw[jttB, densely dotted, line width=0.8pt] ({axis cs:60,0} |- {rel axis cs:0,0}) -- ({axis cs:60,0} |- {rel axis cs:0,1});
\draw[jttB, line width=0.8pt] (axis cs:60,0.0556) circle[radius=3pt];
\end{groupplot}
\end{tikzpicture}
\vspace{-2mm}
\caption{\textbf{Evolution of downstream metrics and flow-matching loss during baseline training} (Kubric-A; lower is better).
Larger models attain their best FVMD early and then memorize, as their validation flow-matching loss (dashed) rises while the training loss (solid) keeps falling; dotted lines mark each model's lowest validation loss (thin: checkpoints; thick: 3-checkpoint moving average).}%
\label{fig:dynamics}
\vspace{-6mm}
\end{figure}
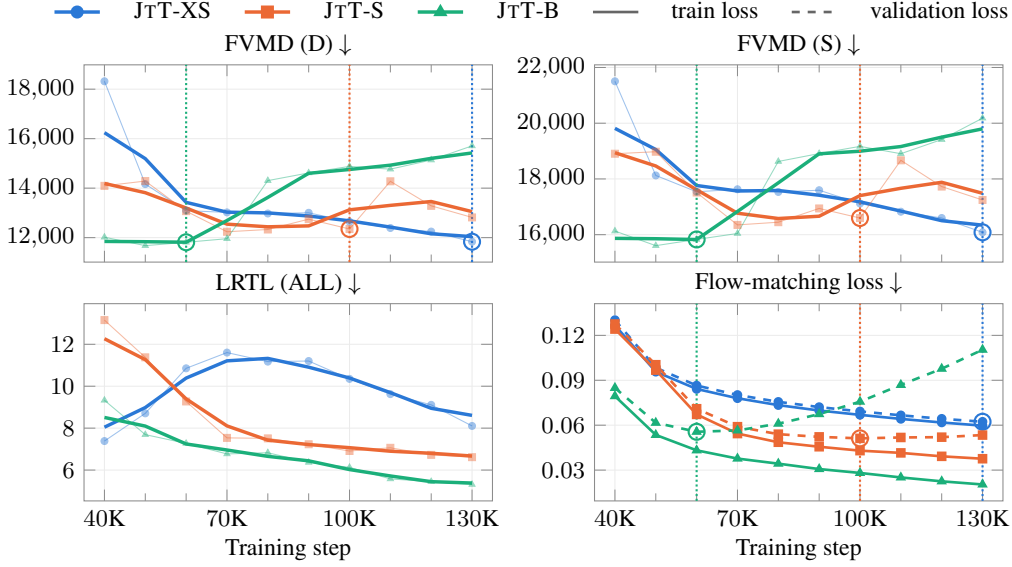

\paragraph{Model size, flow-matching loss, and downstream metrics.}
Prior work on trajectory forecasting typically reports a single checkpoint, sometimes just for one model size~\citep{bharadhwaj24track2act:,thakkar2026forecasting}, without analyzing how metrics evolve during training.
In \cref{fig:dynamics}, we show how model size affects the flow-matching loss and downstream metrics over training for \nickname-XS, -S, and -B (exact values in \cref{tab:scale}).
Larger models converge faster but show clear signs of overfitting: their validation flow-matching loss (on the official Kubric-A validation split, disjoint from the test set; averaged over timesteps 0.1--0.9) rises while the training loss keeps falling; for B, it is lowest at 60K steps and nearly doubles by 130K, and S follows later and more mildly.
The first effect of overfitting is that the model collapses to predicting a single mode for each input image, evidenced by an increase in FVMD, while LRTL (and Best-of-$K$, \cref{fig:dynamics-bok}) keeps improving, with the LRTL of B eventually falling below that of the ground truth (5.31 vs.\ 6.68 at 130K).
XS, in contrast, does not overfit: its training and validation losses stay close, and its FVMD decreases steadily.
Overall, plausibility (LRTL) and diversity (FVMD) do not improve together, and rankings depend on when training stops: at 50K steps, B has the lowest FVMD (D) of the three models; at 130K, the highest.

\paragraph{Validation loss and model selection.}
The validation flow-matching loss is indicative of distributional fidelity: its minimum coincides with the best FVMD (D) and (S) for XS at 130K (and, for FVMD (S), also at 200K when trained twice as long; \cref{tab:scale-xs}); it is one checkpoint later for B (60K vs.\ 50K), a 1.2--1.4\% FVMD difference within the checkpoint-to-checkpoint noise of FVMD, whose moving average is indeed lowest at 60K; and it is later for S (100K vs.\ 70K), whose validation loss is nearly flat from 90K to 120K.
This suggests that the validation loss, which needs one forward pass per sample, could serve as a cheap proxy for FVMD, which requires sampling many futures per image.
To avoid single-checkpoint results, we evaluate an EMA checkpoint every 10K steps and report the average over the last five (60K--100K), together with the best among them, as Average (Best).

\begin{table}[ht]
\centering
\small
\caption{\textbf{Motion generation quality on Kubric-A, following \WHN{}.}
Prior methods: single checkpoint; \nickname: Average (Best) of the last five checkpoints.
\nickname improves over prior methods on FVMD and LRTL, nearing the ground truth; \GPTF{} is best on Best-of-$K$.
$^\dagger$: reimplementation; $^*$: estimated.}%
\label{tab:overall}
\setlength{\tabcolsep}{4pt}
\begin{tabular}{@{}lccccc@{}}
\toprule
\textbf{Model} & \textbf{Size} & \textbf{FVMD (D)} $\downarrow$ & \textbf{FVMD (S)} $\downarrow$ & \textbf{Best-of-$K$} $\downarrow$ & \textbf{LRTL (ALL)} $\downarrow$ \\
\midrule
GT & --- & 4,941 & 8,845 & 110.85 & 6.68 \\
\TA{} (from \WHN{}) & 600M & 16,735 & 22,509 & 250.80 & 15.80 \\
\WHN{} (L, latent-space) & 250M & 13,745 & 17,838 & 127.00 & 14.10 \\
\WHN{} (B, coordinate-space) & 50M & 14,519 & 18,551 & 137.80 & 23.70 \\
\MF{}$^\dagger$ & 100M & 20,178 & 26,379 & 186.47 & 8.20 \\
\GPTF{} & 30M$^*$ & --- & 17,786 & 95.60 & 14.60 \\
\midrule
\multicolumn{6}{l}{\textbf{\nickname}} \\
S (Sparse) & 50M & 12,546 (12,238) & 16,769 (16,351) & 160.25 (129.77) & 7.69 (6.91) \\
B (Sparse) & 200M & 13,514 (11,812) & 17,716 (15,829) & 124.37 (115.24) & 6.67 (6.13) \\
S (Dense) & 55M & 11,807 (11,591) & 15,695 (15,452) & 135.85 (128.69) & 6.84 (6.64) \\
B (Dense) & 230M & 13,266 (12,333) & 17,437 (15,958) & 130.44 (118.76) & 6.80 (6.13) \\
\bottomrule
\end{tabular}
\vspace{-3mm}
\end{table}
\begin{figure}[htb]
\centering
\includegraphics[width=\linewidth]{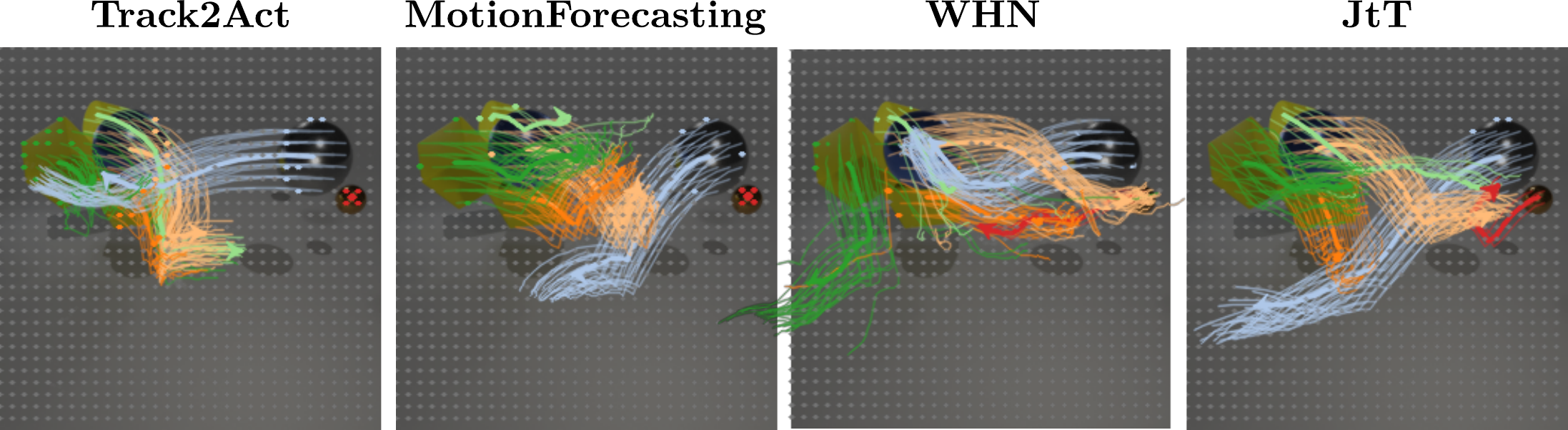}
\caption{\textbf{Qualitative results on Kubric from \WHN{}~\citep{boduljak2026whn}.}
Baselines often miss smaller objects (\hlbox{trackred}{\textcolor{white}{red}}) or object parts (\hlbox{trackdarkgreen}{\textcolor{white}{dark green}}), and frequently fail to capture boundary points accurately (\hlbox{trackpeach}{peach}), primarily due to their low-resolution conditioning.
In contrast, our method reliably captures objects and their parts while achieving near-perfect boundary accuracy.
This highlights our token design, which uses full-resolution RGB at the query points.}%
\label{fig:comparison}
\vspace{-3mm}
\end{figure}

\paragraph{Comparison with prior work.}
\Cref{tab:overall} compares \nickname, without (Sparse) and with the support architecture (Dense), with prior trajectory generators on the benchmark of \WHN{}.
Since checkpoint selection is crucial (\cref{fig:dynamics}), we report both the average over our last five checkpoints and the best among them (all in \cref{tab:scale}), whereas for prior methods we copy their single reported checkpoint; although this favors prior methods, every \nickname variant improves over all of them on FVMD (D), FVMD (S), and rigidity, with LRTL close to the ground truth (6.67--7.69 vs.\ 6.68).
\GPTF{} obtains a better Best-of-$K$, but at the expense of rigidity (LRTL 14.60): its samples are closer to the ground truth pointwise, yet less physically plausible.
Consistent with \cref{tab:diffusion-results}, prior methods that diffuse absolute coordinates (\TA{}, \WHN{}, \GPTF{}) obtain poor rigidity (LRTL 14.10--23.70), whereas \MF{}, the only one diffusing velocities, is the most rigid among them (8.20); \nickname nevertheless outperforms \MF{} on all metrics without using the occlusion labels it requires, highlighting the benefit of full-resolution RGB at the query points.

\paragraph{Baseline details.}
We ablate one design choice at a time, starting from a baseline that corresponds to \cref{fig:method} with every query in the support set (Sparse): the Support Encoder refines all query tokens via self-attention and cross-attention to image patches, and the Velocity Head decodes them directly.
It samples RGB at the query points and uses uniform time sampling (logit-normal in \cref{tab:overall}).

\begin{table}[ht]
\centering
\small
\caption{\textbf{The impact of diffusion target parameterization and timestep scheduling.}
Relative parameterizations outperform absolute coordinates; diffusing velocities with logit-normal timestep sampling gives the best overall trade-off between diversity and rigidity.}%
\label{tab:diffusion-results}
\begin{tabular}{lccccc}
\toprule
Model & FVMD (D) & FVMD (S) & Best-of-$K$ & LRTL (ALL) & LRTL (FG) \\
\midrule
\multicolumn{6}{l}{$\tau \sim \operatorname{Uniform}[0,1]$} \\
\midrule
absolute & 18,695 (15,750) & 23,719 (20,772) & 153.54 (149.67) & 14.76 (11.65) & 13.22 (11.56) \\
velocities & 19,927 (17,785) & 25,914 (23,518) & 125.31 (122.19) & 5.12 (4.81)   & 5.38 (5.17)   \\
offsets  & 13,915 (13,164) & 19,100 (18,245) & 133.68 (124.97) & 9.74 (8.61)   & 9.90 (8.93)   \\
\midrule
\multicolumn{6}{l}{$\tau \sim \operatorname{LogitNormal}(0,1)$} \\
\midrule
absolute & 20,284 (19,140) & 24,678 (23,801) & 125.15 (118.90) & 25.59 (21.87) & 19.25 (17.16) \\
velocities & 14,136 (12,386) & 18,523 (16,589) & 138.27 (122.00) & 6.82 (6.43)   & 6.35 (6.10)   \\
offsets  & 13,883 (13,207) & 17,814 (17,013) & 145.56 (128.79) & 15.91 (14.00) & 13.82 (12.24) \\
\bottomrule
\end{tabular}
\vspace{-3mm}
\end{table}

\paragraph{Parameterization and timestep sampling.}
With all variants normalized isotropically and the rest of the setup fixed, relative parameterizations almost always outperform the absolute one, but only velocities capture rigidity and surpass the previous state of the art on every best-case metric (\cref{tab:diffusion-results}).
With uniform timestep sampling, however, relative parameterizations lose sample diversity; logit-normal sampling alleviates this, so we adopt velocities with logit-normal sampling throughout.

\paragraph{Image conditioning.}
Sampling RGB at full resolution at the query point substantially improves Best-of-$K$ (182.73 $\to$ 138.27) and rigidity (LRTL 8.49 $\to$ 6.82) and resolves the boundary failures of prior methods (\cref{fig:comparison}), whereas DINO features bilinearly sampled at the query point, as in \MF{}, degrade most metrics (\cref{tab:token} in the appendix).
For the image tokens, adding register tokens~\citep{darcet2024vision} to DINOv2 improves all metrics (\eg Best-of-$K$ 138.27 $\to$ 126.00), and multi-layer (DPT) features further improve Best-of-$K$ and foreground rigidity; other extractors do not help consistently (\cref{tab:feature-ablation}), so we use RGB with DINOv2 layers $3, 6, 9, 12$.

\paragraph{Dense support architecture.}
Beyond enabling dense decoding, the support architecture improves FVMD and foreground rigidity for both S and B (details in \cref{tab:support} in the supp.~mat.): for S, FVMD (D) 12,546 $\to$ 11,807 and LRTL (FG) 6.57 $\to$ 5.82; for B, 13,514 $\to$ 13,266 and 6.01 $\to$ 5.87.
For S, it improves every metric; the slight Best-of-$K$ and overall LRTL degradation for B mirrors the overfitting trends of \cref{fig:dynamics} in reverse: we attribute it
to the harder objective of decoding all queries through a randomly re-partitioned support set, which alleviates overfitting.

\begin{table}
\centering
\small
\caption{\textbf{Explicit oracle geometry helps.}
Inverse depth or 3D world coordinates mainly improve Best-of-$K$, while in rigidity our 2D-only model closes most of the gap from prior 2D methods to the 3D oracle.
Kubric-A/D are re-rendered with 2D--3D correspondence, so numbers differ from \cref{tab:overall}.}%
\label{tab:geometry}
\setlength{\tabcolsep}{4.5pt}
\begin{tabular}{lccccc}
\toprule
Model & FVMD (D) & FVMD (S) & Best-of-$K$ & LRTL (ALL) & LRTL (FG) \\
\midrule
\multicolumn{6}{l}{\textit{Kubric-A}} \\
2D & 14,097 (12,080) & 20,074 (17,790) & 120.04 (111.50) & 6.16 (5.75) & 5.92 (5.70) \\
2D + Inverse Depth & 13,946 (12,560) & 19,827 (17,962) & 92.51 (83.98) & 5.59 (4.21) & 5.97 (4.77) \\
3D & 12,909 (12,162) & 18,518 (17,641) & 95.02 (92.37) & 4.37 (3.86) & 4.82 (4.38) \\
\multicolumn{6}{l}{\textit{Kubric-D}} \\
2D & 19,210 (17,848) & 28,412 (25,903) & 94.16 (89.68) & 2.05 (1.95) & 1.69 (1.60) \\
2D + Inverse Depth & 18,652 (17,223) & 27,796 (25,746) & 64.03 (56.80) & 1.81 (1.55) & 1.80 (1.59) \\
3D & 19,500 (18,955) & 29,409 (28,629) & 76.77 (73.22) & 1.95 (1.67) & 1.93 (1.72) \\
\bottomrule
\end{tabular}
\vspace{-3mm}
\end{table}

\paragraph{Does explicit 3D geometry help?}
For these experiments, we re-render Kubric-A and Kubric-D with direct point-to-point correspondence between 2D and 3D trajectories, so that the 2D and 3D variants are trained and evaluated on identical points; absolute numbers therefore differ slightly from \cref{tab:overall}.
Oracle geometry from the simulator, as inverse depth at the query points or queries and trajectories in 3D world coordinates, helps mainly in Best-of-$K$ (\cref{tab:geometry}).
In rigidity, however, our 2D-only model closes most of the gap between prior 2D methods and the 3D oracle (LRTL 6.16 vs.\ 4.37, compared to 8.20--23.70 for prior work; \cref{tab:overall}), verifying the effectiveness of our design.

\section{Conclusion}%
\label{sec:conclusions}

We showed that diverse and plausible rigid-body motion can be forecast from a single image with \nickname, a simple diffusion transformer operating directly in trajectory space, without trajectory VAEs, fixed grids, or rigidity priors, on synthetic multi-future data.
Its key ingredients are diffusing per-frame velocities with logit-normal timestep sampling, full-resolution RGB at the query points, and a randomized support set for arbitrary and dense queries.
As plausibility and diversity do not improve together during training, fair comparison requires checkpoint-averaged evaluation on such data, under which \nickname outperforms prior methods on distributional and rigidity metrics.
We hope our findings promote better evaluation practices in the field; we leave rigorous real-data evaluation to future work.
\section*{Acknowledgements}%
\label{sec:acknowledgements}

The authors acknowledge the use of resources provided by the Isambard-AI National AI Research Resource (AIRR)~\citep{mcintoshsmith2024isambard}.
Isambard-AI is operated by the University of Bristol and is funded by the UK Government's Department for Business, Innovation, Science and Trade (DBIST) via UK Research and Innovation; and the Science and Technology Facilities Council [ST/AIRR/I-A-I/1023].

\section*{AI Use Statement}%
\label{sec:ai_statement}

In this work, we used generative AI tools to polish the paper, improving clarity and rephrasing text, and to identify additional relevant sources, which we read and thoroughly checked manually.
We have \emph{not} used generative AI tools in the implementation of our method, nor for research ideation, experiment design, or synthetic data generation, which were carried out by the authors.
However, we \emph{did} use generative AI to help write visualization code, for instance to output plots and animations for the paper and for debugging.
We have reviewed all AI-assisted work; in particular, we checked the resulting plots and visualizations and ensured that they are correct and faithfully reflect our results.
We take responsibility for the final content of this work, including text, claims, or artifacts produced with the aid of generative AI.


\bibliographystyle{iclr2027/iclr2027_conference}
\bibliography{references,vedaldi_general,vedaldi_specific,yslan_general}

\clearpage
\appendix

\section{Extended related work}%
\label{app:related}

This section expands the discussion of \cref{sec:related} with details omitted for space.

\subsection{Image-conditioned trajectory forecasting}%
\label{sec:related_forecasting}

We forecast object motions as 2D point trajectories from a single image.
Existing methods differ in what they predict (absolute vs.\ relative coordinates) and how they model uncertainty (deterministic vs.\ generative).

\paragraph{Deterministic regression of absolute coordinates.}
ATM~\citep{wen2024anypointtrajectorymodelingpolicy} pseudo-labels LIBERO~\citep{liu2023libero} with CoTracker~\citep{karaev2024cotrackerbettertrack} and trains a deterministic transformer to regress trajectories of $32$ active points; \TraMoE{}~\citep{yang25tra-moe:} adds a mixture-of-experts head.
Regressing a single output per query, these methods cannot represent the multimodal future implied by a single image, and are biased toward active points by training-time supervision.

\paragraph{Generative diffusion of absolute coordinates.}
\TA{}~\citep{bharadhwaj24track2act:} samples $400$ trajectories with a denoising diffusion model~\citep{ho20denoising} conditioned on an image and a goal image, pseudo-labeled from~\citet{goyal17the-something,damen2022RESCALING,brohan2022rt1,walke2023bridgedata}.
While stochastic, it diffuses absolute coordinates and ingests the image as a single global vector with a 600M-parameter network, which in practice hallucinates objects and misses small ones.
A separate, training-free method~\citep{pandey2024motionmodeshappennext} guides a pre-trained image-to-flow model~\citep{shi2024motioni2vconsistentcontrollableimagetovideo} with hand-crafted energies, but only handles one pre-segmented object.

\paragraph{Latent trajectory diffusion.}
These methods generate trajectories in a learned latent space, typically via a VAE~\citep{walker2016uncertainfutureforecastingstatic,li2018flowgroundedspatialtemporalvideoprediction}; see \cref{app:related_latent} for details.
Closest to our setting is \textbf{\WHN{}}~\citep{boduljak2026whn}, which trains a $\beta$-VAE over a fixed $32{\times}32$ trajectory grid and runs rectified flow~\citep{lipman22flow,liu23flow} in the resulting latent space, conditioned on patch-level DINOv2~\citep{oquab2023dinov2} features.
\WHN{} is generative and uses patch-level conditioning, but the VAE introduces \emph{jitter} and \emph{boundary errors}, and the fixed $32{\times}32$ grid couples training and inference patterns, limiting the flexibility required for downstream applications.
\nickname builds on \WHN{}'s generative, image-grounded formulation and rigorous evaluation, but removes the VAE entirely, generates per-frame velocities directly in coordinate space, and supports arbitrary inference patterns at any density.
Following \WHN{}, we evaluate motion plausibility with FVMD~\citep{liu2024fr} and the rigidity-based LRTL score~\citep{karazija24learning}, in the spirit of~\citet{salimans2016improved,unterthiner2018towards,huang2024vbench,bansal25videophy:}, since pixel-level metrics are poor proxies for motion generation~\citep{boduljak2026whn}.

\subsection{Earlier latent-VAE trajectory methods}%
\label{app:related_latent}

\citet{walker2016uncertainfutureforecastingstatic} use a DCT-based linear compression of trajectory offsets together with a VAE to sample futures from static images.
The linear compression has limited expressivity, and the evaluation does not test physical plausibility or distributional coverage.
\citet{li2018flowgroundedspatialtemporalvideoprediction} use a VAE to generate optical flow maps from a static image and translate them to RGB; this inherits the failure modes of frame-pair flow (drift under occlusion, no long-range consistency) and uses pixel-level metrics rather than motion-aware ones.
Both serve as early demonstrations that latent generative modeling of motion is viable, but neither addresses the multi-object, rigidity-aware setting we target.

\subsection{Recent work in detail}%
\label{app:related_recent}

\paragraph{\TL{}.}
\TL{}~\citep{zhang2026trajloomdensefuturetrajectory} extends the latent-VAE formulation of \WHN{} to \emph{video} conditioning at larger scale: it diffuses VAE latents on a fixed grid using Wan2.1~\citep{wan2025} RGB video latents and trains with auxiliary VAE losses and rollout fine-tuning.
The contribution is essentially an offset parameterization relative to the grid, plus additional video signal at larger scale.
The underlying recipe, latent diffusion over a fixed grid, and its limitations (no arbitrary query patterns, no density beyond the training grid) carry over, while the offset-based parameterization is suboptimal, as we show in this work.

\paragraph{\MF{}.}
\MF{}~\citep{thakkar2026forecasting} (``Forecasting Motion in the Wild'') extends \TA{} to in-the-wild forecasting on \emph{wildlife} footage and is, architecturally, \TA{} with the goal image dropped and ResNet-50 swapped for DINOv3~\citep{simeoni2025dinov3}.
The model is a joint-attention transformer over query points, which scales quadratically in the number of queries and cannot run densely.
The relative-coordinate parameterization is adopted but never ablated, so its importance is not established.
The evaluation runs \WHN{} \emph{zero-shot} on animal videos, even though \WHN{} is trained only on Kubric, so it remains unclear whether the proposed method outperforms \WHN{} under matched supervision.
Even in this zero-shot setting, \WHN{} still outperforms \MF{} on the acceleration-based distributional metrics.
The model also inherits the low-resolution image conditioning of \TA{}, and consequently the same failure mode at object boundaries.
The setting also differs from ours: a single animal moving with simple, mostly translational motion, rather than scenes with multiple interacting objects and rigid-body collisions.
In addition, \MF{} requires occlusion labels as input, whereas we deliberately minimize the number of inputs.
Finally, \cref{tab:token} shows that, within our architecture and on the benchmark of~\citet{boduljak2026whn}, bilinearly sampling DINO features at the query points, as in \MF{}, is worse than sampling full-resolution RGB on every metric except foreground rigidity.

\paragraph{Reimplementation of \MF{}.}
Neither the implementation nor the dataset of~\citet{thakkar2026forecasting} is publicly available, so we reimplement the method ourselves.
We start from the \TA{} codebase and strictly follow the architecture and training recipe described in the paper.
We set the trajectory horizon to $T=24$ Kubric timesteps, matching prior work and our method, and use ground-truth occlusion labels and the same isotropic delta normalization as for our method.
We train for 140 epochs, as in the paper, which amounts to about 180K steps.
In contrast, the checkpoints we report for our models are trained for at most 100K steps, since larger models overfit beyond that point (\cref{fig:dynamics}).

\paragraph{Parameter count of \GPTF{}.}%
\label{app:gptf-params}
No public implementation of~\citet{lu2026generative} is available, so we estimate its size from the architecture described in the paper.
On Kubric, the model is a DiT with 6 blocks of width $d=384$; each block has temporal, spatial, and point--image attention ($3 \times 4d^2$), an MLP, which we assume has ratio 4 ($8d^2$), and AdaLN modulation for its four sublayers ($12d^2$), giving about $32d^2 \approx 4.7$M parameters per block and about 28M for all six.
The feature projection, upsampling layer, embeddings, and output layer add about 2M, for about 30M in total.
The same estimate for the Kinetics model (12 blocks of width 768) gives about 230M.
As for \nickname{}, we exclude the pre-trained image encoder (DINOv3-S).

\section{More results}%
\label{app:results}

\Cref{tab:token,tab:support,tab:feature-ablation} report the full conditioning, support-architecture, and feature-extractor ablations, and \cref{tab:checkpoints} the per-checkpoint results behind \cref{tab:overall}.

\begin{table}[htb]
\centering
\small

\caption{\textbf{Impact of full-resolution inputs on motion generation.}
RGB alone yields strong results, yet more explicit representations, such as segmentation and inverse depth, further improve performance.}%
\label{tab:token}
\begin{tabular}{@{} l rrrrr @{}}
\toprule
Token with & \multicolumn{1}{c}{FVMD (D)} & \multicolumn{1}{c}{FVMD (S)} & \multicolumn{1}{c}{Best-of-$K$} & \multicolumn{1}{c}{LRTL (ALL)} & \multicolumn{1}{c}{LRTL (FG)} \\
\midrule
Nothing           & 13,458 (13,006) & 17,662 (17,147) & 182.73 (175.60) & 8.49 (7.61) & 5.95 (5.56) \\
DINO             & 14,470 (13,901) & 19,189 (18,357) & 184.19 (170.87) & 8.94 (8.48) & 6.15 (5.92) \\
\midrule
RGB            & 14,136 (12,386) & 18,523 (16,589) & 138.27 (122.00) & 6.82 (6.43) & 6.35 (6.10) \\
Inverse Depth       & 15,572 (13,653) & 20,116 (17,716) & 88.72 (81.63)  & 4.11 (3.82) & 4.56 (4.30) \\
Segm (binary)   & 13,232 (12,753) & 17,281 (16,816) & 93.49 (86.18)  & 5.84 (5.29) & 6.58 (5.99) \\
Segm (instance) & 13,356 (12,888) & 17,666 (17,205) & 104.64 (97.77)  & 6.01 (5.24) & 6.77 (5.93) \\
\bottomrule
\end{tabular}
\end{table}
\begin{table}
\centering
\small
\caption{\textbf{Adding a support-point architecture.}
Besides enabling dense decoding, the support-point architecture (Dense) improves FVMD and foreground rigidity (LRTL (FG)) over the baseline (Sparse) for both model sizes, and improves every metric for S.
The slight degradation in Best-of-$K$ and LRTL (ALL) for B is expected, as the Dense model solves a harder task, and is not visible qualitatively.}%
\label{tab:support}
\begin{tabular}{lrrrrr}
\toprule
Model & FVMD (D) & FVMD (S) & Best-of-$K$ & LRTL (ALL) & LRTL (FG) \\
\midrule
B & 13,514 (11,812) & 17,716 (15,829) & 124.37 (115.24) & 6.67 (6.13) & 6.01 (5.58) \\
B (Dense) & 13,266 (12,333) & 17,437 (15,958) & 130.44 (118.76) & 6.80 (6.13) & 5.87 (5.31) \\
S & 12,546 (12,238) & 16,769 (16,351) & 160.25 (129.77) & 7.69 (6.91) & 6.57 (6.26) \\
S (Dense) & 11,807 (11,591) & 15,695 (15,452) & 135.85 (128.69) & 6.84 (6.64) & 5.82 (5.46) \\
\bottomrule
\end{tabular}
\end{table}
\begin{table}[h]
\centering
\small
\caption{\textbf{The impact of the image feature extractor on motion quality} on Kubric-A.
All models diffuse velocities; only the feature extractor changes.
Adding register tokens (Reg) improves all metrics, and multi-layer (DPT) features further improve Best-of-$K$ and foreground rigidity.
Larger or alternatively pretrained extractors (DINOv2 G, 4RC, VGGT-Omega, VJEPA2.1) do not yield consistent gains.
Results are reported as Average (Best).}%
\label{tab:feature-ablation}
\setlength{\tabcolsep}{4.5pt}
\begin{tabular}{@{}lccccc@{}}
\toprule
Feature extractor & FVMD (D) & FVMD (S) & Best-of-$K$ & LRTL (ALL) & LRTL (FG) \\
\midrule
DINOv2 S & 14,136 (12,386) & 18,523 (16,589) & 138.27 (122.00) & 6.82 (6.43) & 6.35 (6.10) \\
DINOv2 S+Reg & 13,441 (12,793) & 17,623 (16,763) & 126.00 (116.75) & 6.67 (5.95) & 6.21 (5.86) \\
DINOv2 S+Reg+DPT & 13,514 (11,812) & 17,716 (15,829) & 124.37 (115.24) & 6.67 (6.13) & 6.01 (5.58) \\
DINOv2 G+Reg+DPT & 14,519 (12,522) & 18,482 (16,442) & 123.72 (104.64) & 6.75 (5.88) & 6.04 (5.57) \\
\midrule
4RC+DPT & 13,363 (12,227) & 17,441 (16,284) & 128.20 (121.22) & 6.40 (5.56) & 5.88 (5.29) \\
VGGT-Omega+DPT & 13,821 (12,824) & 18,045 (17,054) & 150.18 (125.42) & 7.27 (6.43) & 6.53 (6.09) \\
VJEPA2.1+DPT & 13,988 (12,257) & 18,274 (16,416) & 146.00 (131.63) & 6.71 (5.95) & 6.02 (5.65) \\
\bottomrule
\end{tabular}
\end{table}

\paragraph{Sampler.}
\Cref{tab:sampling} compares our default 1000-step SDE (Euler--Maruyama) sampler with Euler integration of the probability-flow ODE.
The SDE sampler needs at least 250 steps to be competitive, whereas the ODE sampler is robust to the number of steps: with only 10 steps, it improves both FVMD variants and Best-of-$K$ while using $100\times$ fewer function evaluations, at the cost of slightly lower rigidity (LRTL 7.05 vs.\ 6.77).

\begin{table}[!ht]
\centering
\small
\renewcommand{\arraystretch}{0.92}
\caption{\textbf{Effect of the sampler and number of function evaluations (NFE)} on Kubric-A, for a single \nickname-B (Sparse) checkpoint at 70K steps.
Our default is the 1000-step SDE (Euler--Maruyama) sampler.}%
\label{tab:sampling}
\begin{tabular}{@{}lrrrrrr@{}}
\toprule
Sampler & NFE & FVMD (D) & FVMD (S) & Best-of-$K$ & LRTL (ALL) & LRTL (FG) \\
\midrule
\multirow{6}{*}{ODE} & 10 & 11,614 & 15,263 & 115.59 & 7.05 & 6.41 \\
& 25 & 11,404 & 15,221 & 126.82 & 7.95 & 7.18 \\
& 50 & 11,365 & 15,266 & 133.83 & 8.35 & 7.51 \\
& 100 & 11,357 & 15,311 & 138.16 & 8.59 & 7.71 \\
& 250 & 11,359 & 15,348 & 141.45 & 8.74 & 7.84 \\
& 1000 & 11,362 & 15,370 & 143.29 & 8.83 & 7.90 \\
\midrule
\multirow{6}{*}{SDE} & 10 & 32,396 & 39,223 & 2,797.51 & 50.47 & 31.02 \\
& 25 & 21,258 & 26,333 & 337.13 & 16.87 & 13.49 \\
& 50 & 17,487 & 22,372 & 260.94 & 12.45 & 10.95 \\
& 100 & 14,070 & 18,700 & 186.54 & 9.78 & 8.87 \\
& 250 & 11,901 & 16,122 & 139.93 & 7.65 & 6.93 \\
& 1000 & 11,957 & 16,043 & 126.12 & 6.77 & 6.19 \\
\bottomrule
\end{tabular}
\end{table}

\begin{figure}[t]
\centering
\begin{tikzpicture}
\begin{groupplot}[
group style={group size=1 by 2, vertical sep=2.0cm},
width=0.75\textwidth, height=0.4\textwidth,
xmin=35, xmax=135, xtick={40,70,100,130}, minor xtick={50,60,80,90,110,120}, xticklabel={\pgfmathprintnumber{\tick}K},
xlabel={Training step}, xlabel style={font=\footnotesize, yshift=2pt},
tick label style={font=\footnotesize}, title style={font=\small, yshift=-4pt},
scaled y ticks=false, yticklabel style={/pgf/number format/.cd, fixed, 1000 sep={\mathord{,}}},
axis line style={black!45}, tick style={black!45}, grid=major, grid style={black!8, line width=0.3pt},
legend style={font=\small, draw=none, fill=none, legend columns=1, row sep=2pt, legend cell align=left, at={(1.03,0.5)}, anchor=west},
]
\nextgroupplot[title={FVMD (D) $\downarrow$}]
\addplot[jttXS, line width=0.4pt, opacity=0.45, mark=*, mark size=1.3pt, forget plot] coordinates {(40,18319.062) (50,14164.582) (60,13091.881) (70,13019.896) (80,12975.826) (90,13004.096) (100,12643.575) (110,12389.515) (120,12244.834) (130,11835.232)};
\addplot[jttXS, line width=1.3pt, mark=none, legend image post style={mark=*, mark size=1.8pt}] coordinates {(40,16241.822) (50,15191.842) (60,13425.453) (70,13029.201) (80,12999.939) (90,12874.499) (100,12679.062) (110,12425.975) (120,12156.527) (130,12040.033)};
\addlegendentry{\nickname-XS}
\addplot[jttS, line width=0.4pt, opacity=0.45, mark=square*, mark size=1.3pt, forget plot] coordinates {(40,14093.199) (50,14286.340) (60,13068.819) (70,12237.690) (80,12328.595) (90,12746.655) (100,12349.562) (110,14276.285) (120,13286.094) (130,12817.273)};
\addplot[jttS, line width=1.3pt, mark=none, legend image post style={mark=square*, mark size=1.8pt}] coordinates {(40,14189.770) (50,13816.119) (60,13197.616) (70,12545.035) (80,12437.647) (90,12474.937) (100,13124.167) (110,13303.980) (120,13459.884) (130,13051.683)};
\addlegendentry{\nickname-S}
\addplot[jttB, line width=0.4pt, opacity=0.45, mark=triangle*, mark size=1.3pt, forget plot] coordinates {(40,12017.968) (50,11675.977) (60,11812.407) (70,11957.168) (80,14304.908) (90,14615.274) (100,14880.139) (110,14767.617) (120,15136.429) (130,15708.267)};
\addplot[jttB, line width=1.3pt, mark=none, legend image post style={mark=triangle*, mark size=1.8pt}] coordinates {(40,11846.973) (50,11835.451) (60,11815.184) (70,12691.494) (80,13625.783) (90,14600.107) (100,14754.343) (110,14928.062) (120,15204.104) (130,15422.348)};
\addlegendentry{\nickname-B}
\nextgroupplot[title={FVMD (S) $\downarrow$}]
\addplot[jttXS, line width=0.4pt, opacity=0.45, mark=*, mark size=1.3pt, forget plot] coordinates {(40,21504.536) (50,18125.482) (60,17539.278) (70,17629.147) (80,17532.914) (90,17595.794) (100,17117.613) (110,16824.290) (120,16598.983) (130,16084.196)};
\addplot[jttXS, line width=1.3pt, mark=none, legend image post style={mark=*, mark size=1.8pt}] coordinates {(40,19815.009) (50,19056.432) (60,17764.636) (70,17567.113) (80,17585.952) (90,17415.440) (100,17179.232) (110,16846.962) (120,16502.490) (130,16341.589)};
\addlegendentry{\nickname-XS}
\addplot[jttS, line width=0.4pt, opacity=0.45, mark=square*, mark size=1.3pt, forget plot] coordinates {(40,18905.157) (50,18976.823) (60,17512.305) (70,16350.717) (80,16447.447) (90,16941.527) (100,16597.454) (110,18668.485) (120,17726.122) (130,17244.561)};
\addplot[jttS, line width=1.3pt, mark=none, legend image post style={mark=square*, mark size=1.8pt}] coordinates {(40,18940.990) (50,18464.762) (60,17613.282) (70,16770.156) (80,16579.897) (90,16662.143) (100,17402.489) (110,17664.020) (120,17879.722) (130,17485.341)};
\addlegendentry{\nickname-S}
\addplot[jttB, line width=0.4pt, opacity=0.45, mark=triangle*, mark size=1.3pt, forget plot] coordinates {(40,16129.730) (50,15606.083) (60,15828.663) (70,16042.628) (80,18627.156) (90,18911.210) (100,19169.120) (110,18905.608) (120,19413.616) (130,20181.348)};
\addplot[jttB, line width=1.3pt, mark=none, legend image post style={mark=triangle*, mark size=1.8pt}] coordinates {(40,15867.907) (50,15854.826) (60,15825.791) (70,16832.815) (80,17860.331) (90,18902.495) (100,18995.313) (110,19162.782) (120,19500.191) (130,19797.482)};
\addlegendentry{\nickname-B}
\end{groupplot}
\end{tikzpicture}
\caption{\textbf{Distributional fidelity during training} (Kubric-A; lower is better), enlarged from \cref{fig:dynamics}.
B and S reach their best FVMD at 50K and 70K steps, respectively, and then degrade as they memorize the single ground-truth future per training image, whereas the FVMD of XS decreases steadily.
Thin lines: evaluated checkpoints; thick lines: 3-checkpoint moving average.}%
\label{fig:dynamics-fvmd}
\end{figure}
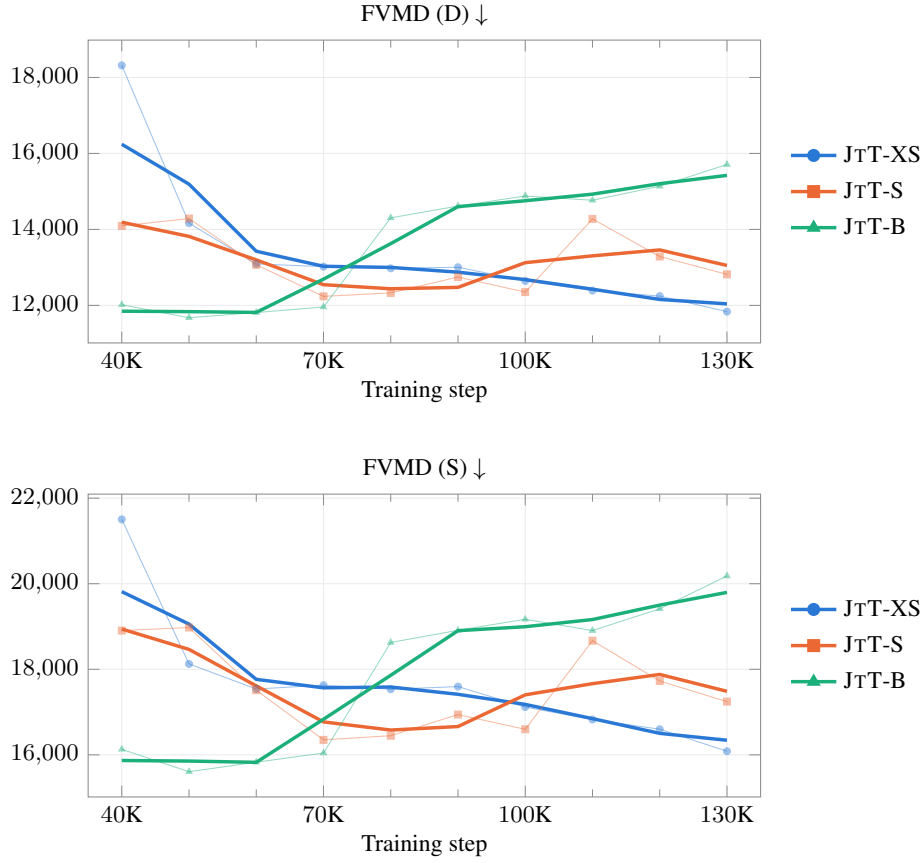

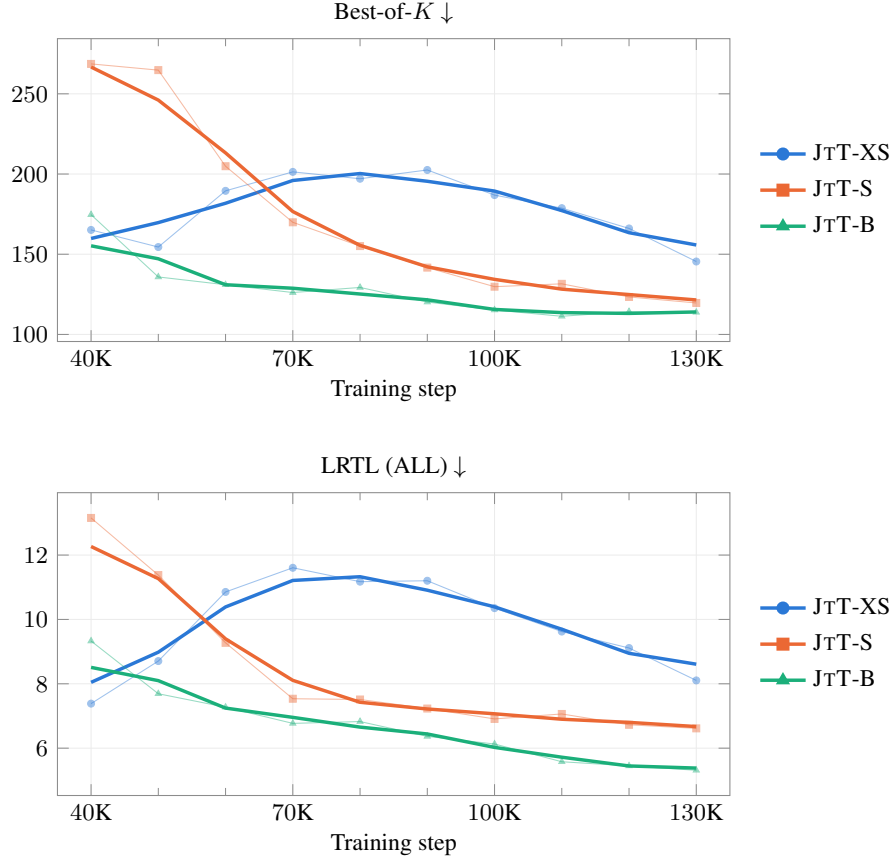
\begin{figure}[t]
\centering
\begin{tikzpicture}
\begin{groupplot}[
group style={group size=1 by 2, vertical sep=2.0cm},
width=0.75\textwidth, height=0.4\textwidth,
xmin=35, xmax=135, xtick={40,70,100,130}, minor xtick={50,60,80,90,110,120}, xticklabel={\pgfmathprintnumber{\tick}K},
xlabel={Training step}, xlabel style={font=\footnotesize, yshift=2pt},
tick label style={font=\footnotesize}, title style={font=\small, yshift=-4pt},
scaled y ticks=false, yticklabel style={/pgf/number format/.cd, fixed, 1000 sep={\mathord{,}}},
axis line style={black!45}, tick style={black!45}, grid=major, grid style={black!8, line width=0.3pt},
legend style={font=\small, draw=none, fill=none, legend columns=1, row sep=2pt, legend cell align=left, at={(1.03,0.5)}, anchor=west},
]
\nextgroupplot[title={Best-of-$K$ $\downarrow$}]
\addplot[jttXS, line width=0.4pt, opacity=0.45, mark=*, mark size=1.3pt, forget plot] coordinates {(40,165.155) (50,154.480) (60,189.524) (70,201.339) (80,197.066) (90,202.533) (100,186.862) (110,178.737) (120,166.084) (130,145.462)};
\addplot[jttXS, line width=1.3pt, mark=none, legend image post style={mark=*, mark size=1.8pt}] coordinates {(40,159.817) (50,169.720) (60,181.781) (70,195.976) (80,200.312) (90,195.487) (100,189.377) (110,177.228) (120,163.428) (130,155.773)};
\addlegendentry{\nickname-XS}
\addplot[jttS, line width=0.4pt, opacity=0.45, mark=square*, mark size=1.3pt, forget plot] coordinates {(40,268.656) (50,264.774) (60,204.860) (70,169.903) (80,155.073) (90,141.659) (100,129.772) (110,131.557) (120,123.368) (130,119.629)};
\addplot[jttS, line width=1.3pt, mark=none, legend image post style={mark=square*, mark size=1.8pt}] coordinates {(40,266.715) (50,246.096) (60,213.179) (70,176.612) (80,155.545) (90,142.168) (100,134.329) (110,128.232) (120,124.851) (130,121.499)};
\addlegendentry{\nickname-S}
\addplot[jttB, line width=0.4pt, opacity=0.45, mark=triangle*, mark size=1.3pt, forget plot] coordinates {(40,174.656) (50,135.836) (60,130.984) (70,126.122) (80,129.298) (90,120.185) (100,115.243) (110,111.326) (120,114.219) (130,113.804)};
\addplot[jttB, line width=1.3pt, mark=none, legend image post style={mark=triangle*, mark size=1.8pt}] coordinates {(40,155.246) (50,147.159) (60,130.980) (70,128.801) (80,125.202) (90,121.575) (100,115.584) (110,113.596) (120,113.116) (130,114.012)};
\addlegendentry{\nickname-B}
\nextgroupplot[title={LRTL (ALL) $\downarrow$}]
\addplot[jttXS, line width=0.4pt, opacity=0.45, mark=*, mark size=1.3pt, forget plot] coordinates {(40,7.384) (50,8.706) (60,10.854) (70,11.605) (80,11.173) (90,11.202) (100,10.350) (110,9.623) (120,9.112) (130,8.105)};
\addplot[jttXS, line width=1.3pt, mark=none, legend image post style={mark=*, mark size=1.8pt}] coordinates {(40,8.045) (50,8.982) (60,10.388) (70,11.211) (80,11.327) (90,10.908) (100,10.392) (110,9.695) (120,8.947) (130,8.609)};
\addlegendentry{\nickname-XS}
\addplot[jttS, line width=0.4pt, opacity=0.45, mark=square*, mark size=1.3pt, forget plot] coordinates {(40,13.154) (50,11.378) (60,9.273) (70,7.536) (80,7.512) (90,7.231) (100,6.909) (110,7.063) (120,6.726) (130,6.615)};
\addplot[jttS, line width=1.3pt, mark=none, legend image post style={mark=square*, mark size=1.8pt}] coordinates {(40,12.266) (50,11.268) (60,9.396) (70,8.107) (80,7.426) (90,7.217) (100,7.068) (110,6.899) (120,6.801) (130,6.671)};
\addlegendentry{\nickname-S}
\addplot[jttB, line width=0.4pt, opacity=0.45, mark=triangle*, mark size=1.3pt, forget plot] coordinates {(40,9.329) (50,7.691) (60,7.273) (70,6.770) (80,6.827) (90,6.366) (100,6.129) (110,5.578) (120,5.449) (130,5.313)};
\addplot[jttB, line width=1.3pt, mark=none, legend image post style={mark=triangle*, mark size=1.8pt}] coordinates {(40,8.510) (50,8.098) (60,7.245) (70,6.957) (80,6.654) (90,6.441) (100,6.024) (110,5.719) (120,5.447) (130,5.381)};
\addlegendentry{\nickname-B}
\end{groupplot}
\end{tikzpicture}
\caption{\textbf{Best-of-$K$ and rigidity during training} (Kubric-A; lower is better), complementing \cref{fig:dynamics}.
Best-of-$K$ and LRTL keep improving for S and B even after their FVMD degrades, as their samples collapse toward the single ground-truth future per training image.
Thin lines: evaluated checkpoints; thick lines: 3-checkpoint moving average.}%
\label{fig:dynamics-bok}
\end{figure}

\begin{figure}[t]
\centering
\begin{tikzpicture}
\begin{axis}[
width=0.75\textwidth, height=0.4\textwidth,
title={Flow-matching loss $\downarrow$}, xlabel={Training step}, xlabel style={font=\footnotesize, yshift=2pt},
xmin=5, xmax=205, xtick={10,50,100,150,200}, minor xtick={20,30,40,60,70,80,90,110,120,130,140,160,170,180,190}, xticklabel={\pgfmathprintnumber{\tick}K},
ymode=log, log ticks with fixed point, ytick={0.02,0.05,0.1,0.2,0.5,1}, yticklabel style={/pgf/number format/.cd, fixed, precision=2},
tick label style={font=\footnotesize}, title style={font=\small, yshift=-4pt},
axis line style={black!45}, tick style={black!45}, grid=major, grid style={black!8, line width=0.3pt},
legend style={font=\small, draw=none, fill=none, legend columns=1, row sep=2pt, legend cell align=left, at={(1.03,0.5)}, anchor=west},
]
\addplot[jttXS, line width=1.0pt, mark=*, mark size=1.3pt] coordinates {(10,1.3640) (20,0.5801) (30,0.2664) (40,0.1267) (50,0.0957) (60,0.0842) (70,0.0780) (80,0.0734) (90,0.0698) (100,0.0669) (110,0.0642) (120,0.0618) (130,0.0597) (140,0.0580) (150,0.0566) (160,0.0554) (170,0.0543) (180,0.0534) (190,0.0526) (200,0.0519)};
\addlegendentry{\nickname-XS (train)}
\addplot[jttXS, line width=1.0pt, dashed, mark=*, mark size=1.3pt, mark options={solid}] coordinates {(10,1.3908) (20,0.5892) (30,0.2722) (40,0.1302) (50,0.0984) (60,0.0865) (70,0.0802) (80,0.0756) (90,0.0721) (100,0.0692) (110,0.0666) (120,0.0643) (130,0.0623) (140,0.0607) (150,0.0594) (160,0.0583) (170,0.0573) (180,0.0565) (190,0.0557) (200,0.0551)};
\addlegendentry{\nickname-XS (val.)}
\addplot[jttS, line width=1.0pt, mark=square*, mark size=1.3pt] coordinates {(10,1.1163) (20,0.2270) (30,0.1164) (40,0.1242) (50,0.0968) (60,0.0672) (70,0.0544) (80,0.0486) (90,0.0456) (100,0.0430) (110,0.0415) (120,0.0392) (130,0.0376) (140,0.0360) (150,0.0346) (160,0.0333) (170,0.0320) (180,0.0307) (190,0.0299) (200,0.0285)};
\addlegendentry{\nickname-S (train)}
\addplot[jttS, line width=1.0pt, dashed, mark=square*, mark size=1.3pt, mark options={solid}] coordinates {(10,1.1389) (20,0.2312) (30,0.1195) (40,0.1276) (50,0.1004) (60,0.0710) (70,0.0590) (80,0.0540) (90,0.0522) (100,0.0512) (110,0.0518) (120,0.0520) (130,0.0534) (140,0.0553) (150,0.0579) (160,0.0612) (170,0.0645) (180,0.0688) (190,0.0736) (200,0.0789)};
\addlegendentry{\nickname-S (val.)}
\addplot[jttB, line width=1.0pt, mark=triangle*, mark size=1.3pt] coordinates {(10,0.8127) (20,0.2375) (30,0.1583) (40,0.0794) (50,0.0535) (60,0.0432) (70,0.0377) (80,0.0343) (90,0.0307) (100,0.0281) (110,0.0251) (120,0.0225) (130,0.0204) (140,0.0186) (150,0.0174) (160,0.0159) (170,0.0148) (180,0.0141) (190,0.0132) (200,0.0129)};
\addlegendentry{\nickname-B (train)}
\addplot[jttB, line width=1.0pt, dashed, mark=triangle*, mark size=1.3pt, mark options={solid}] coordinates {(10,0.8316) (20,0.2422) (30,0.1635) (40,0.0849) (50,0.0616) (60,0.0556) (70,0.0565) (80,0.0610) (90,0.0674) (100,0.0757) (110,0.0869) (120,0.0977) (130,0.1104) (140,0.1242) (150,0.1366) (160,0.1513) (170,0.1640) (180,0.1772) (190,0.1897) (200,0.2005)};
\addlegendentry{\nickname-B (val.)}
\end{axis}
\end{tikzpicture}
\caption{\textbf{Flow-matching loss over the full training run} (Kubric-A; log scale), averaged over diffusion timesteps 0.1--0.9, complementing \cref{fig:dynamics}.
Solid: training loss; dashed: validation loss.
The validation loss of B is lowest at 60K steps and of S at 100K steps, after which both rise while their training losses keep falling; for XS, the two losses stay close and still decrease at 200K steps.}%
\label{fig:velocity-loss-full}
\end{figure}
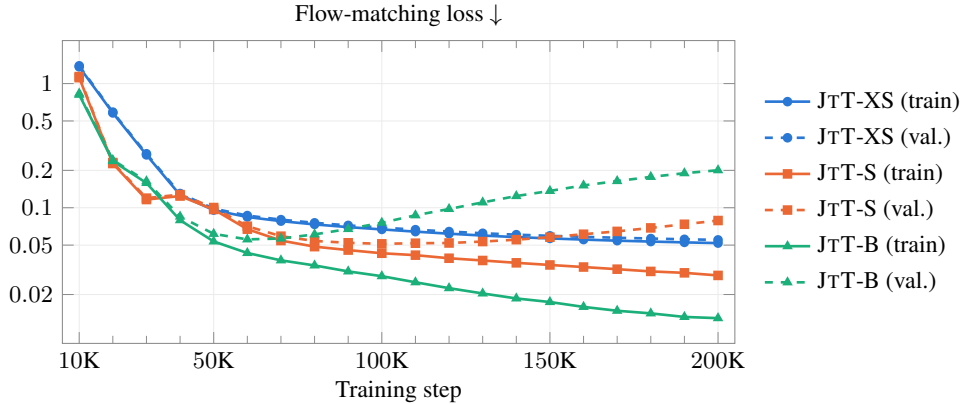

\paragraph{Scale.}
\Cref{tab:scale} compares \nickname-XS, \nickname-S, and \nickname-B, evaluated every 10K steps up to 130K steps.
These trends are easier to see in \cref{fig:dynamics}, enlarged in \cref{fig:dynamics-fvmd,fig:dynamics-bok}; \cref{fig:velocity-loss-full} shows the flow-matching loss over the full 200K steps.
Larger models converge faster: B reaches its lowest FVMD at 50K steps and S at 70K steps.
Afterwards, they start to memorize the single ground-truth future available per training image instead of learning the distribution of futures: qualitatively, their samples collapse toward one future per image.
This is also visible in the metrics, as their FVMD increases while Best-of-$K$ and LRTL continue to improve.
In contrast, the FVMD of XS decreases almost monotonically throughout training.
Since sampling and evaluation are expensive, we did not sample and evaluate S and B beyond 130K steps, as they had clearly started to memorize by then; their flow-matching loss over the full 200K steps is shown in \cref{fig:velocity-loss-full}.
We evaluated XS further, up to 200K steps (\cref{tab:scale-xs}): it keeps improving and reaches FVMD (D) 11,771, Best-of-$K$ 120.78, and LRTL 6.52, comparable to our larger models (\cref{tab:overall}).
Hence, optimizing plausibility (LRTL) does not necessarily optimize diversity: a model can produce increasingly rigid trajectories while covering fewer of the possible futures, as reflected in its FVMD.
This makes model selection tricky, since no single checkpoint is best on all metrics, and it has direct consequences for evaluation: depending on model size and training time, the ranking of models can be reversed.
For example, at 50K steps, B has the lowest FVMD (D) of the three models (11,676 vs.\ 14,165 for XS), whereas at 130K steps, it has the highest (15,708 vs.\ 11,835 for XS).
Prior work largely leaves this model selection implicit, typically reporting a single checkpoint without discussing how it was chosen, which makes it difficult to know which model is better on which metric.
To avoid results that depend on a single, arbitrarily chosen checkpoint, we therefore report the average over the last five checkpoints up to 100K steps, together with the best among them (\cref{sec:experiments}).

\begin{table}[!ht]
\centering
\small
\caption{\textbf{Model performance through training for different model sizes} on Kubric-A, up to 130K steps.
Larger models reach their best FVMD earlier (B at 50K, S at 70K steps), after which FVMD increases while Best-of-$K$ and LRTL keep improving; the FVMD of XS decreases almost monotonically.}%
\label{tab:scale}
\begin{tabular}{@{}lrrrrrr@{}}
\toprule
Model & Step & FVMD (D) & FVMD (S) & Best-of-$K$ & LRTL (ALL) & LRTL (FG) \\
\midrule
\multirow{10}{*}{XS} & 40K & 18,319 & 21,505 & 165.15 & 7.38 & 5.76 \\
& 50K & 14,165 & 18,125 & 154.48 & 8.71 & 7.28 \\
& 60K & 13,092 & 17,539 & 189.52 & 10.85 & 9.36 \\
& 70K & 13,020 & 17,629 & 201.34 & 11.60 & 10.16 \\
& 80K & 12,976 & 17,533 & 197.07 & 11.17 & 9.97 \\
& 90K & 13,004 & 17,596 & 202.53 & 11.20 & 9.97 \\
& 100K & 12,644 & 17,118 & 186.86 & 10.35 & 9.36 \\
& 110K & 12,390 & 16,824 & 178.74 & 9.62 & 8.85 \\
& 120K & 12,245 & 16,599 & 166.08 & 9.11 & 8.49 \\
& 130K & 11,835 & 16,084 & 145.46 & 8.11 & 7.68 \\
\midrule
\multirow{10}{*}{S} & 40K & 14,093 & 18,905 & 268.66 & 13.15 & 9.12 \\
& 50K & 14,286 & 18,977 & 264.77 & 11.38 & 8.37 \\
& 60K & 13,069 & 17,512 & 204.86 & 9.27 & 7.41 \\
& 70K & 12,238 & 16,351 & 169.90 & 7.54 & 6.26 \\
& 80K & 12,329 & 16,447 & 155.07 & 7.51 & 6.46 \\
& 90K & 12,747 & 16,942 & 141.66 & 7.23 & 6.45 \\
& 100K & 12,350 & 16,598 & 129.77 & 6.91 & 6.30 \\
& 110K & 14,276 & 18,668 & 131.56 & 7.06 & 6.31 \\
& 120K & 13,286 & 17,726 & 123.37 & 6.73 & 6.32 \\
& 130K & 12,817 & 17,245 & 119.63 & 6.61 & 6.18 \\
\midrule
\multirow{10}{*}{B} & 40K & 12,018 & 16,130 & 174.66 & 9.33 & 6.91 \\
& 50K & 11,676 & 15,606 & 135.84 & 7.69 & 6.43 \\
& 60K & 11,812 & 15,829 & 130.98 & 7.27 & 6.50 \\
& 70K & 11,957 & 16,043 & 126.12 & 6.77 & 6.19 \\
& 80K & 14,305 & 18,627 & 129.30 & 6.83 & 5.95 \\
& 90K & 14,615 & 18,911 & 120.19 & 6.37 & 5.82 \\
& 100K & 14,880 & 19,169 & 115.24 & 6.13 & 5.58 \\
& 110K & 14,768 & 18,906 & 111.33 & 5.58 & 5.34 \\
& 120K & 15,136 & 19,414 & 114.22 & 5.45 & 5.37 \\
& 130K & 15,708 & 20,181 & 113.80 & 5.31 & 5.30 \\
\bottomrule
\end{tabular}
\end{table}

\begin{table}[!ht]
\centering
\small
\caption{\textbf{Convergence of \nickname-XS} (under 10M parameters) trained for 200K steps on Kubric-A.}%
\label{tab:scale-xs}
\begin{tabular}{@{}lrrrrrr@{}}
\toprule
Model & Step & FVMD (D) & FVMD (S) & Best-of-$K$ & LRTL (ALL) & LRTL (FG) \\
\midrule
\multirow{4}{*}{XS} & 50K & 14,165 & 18,125 & 154.48 & 8.71 & 7.28 \\
& 100K & 12,644 & 17,118 & 186.86 & 10.35 & 9.36 \\
& 150K & 11,737 & 15,873 & 130.47 & 7.21 & 7.12 \\
& 200K & 11,771 & 15,773 & 120.78 & 6.52 & 6.68 \\
\bottomrule
\end{tabular}
\end{table}

\begin{table}[h]
\centering
\small
\caption{\textbf{Per-checkpoint results of the Sparse and Dense models} on Kubric-A, from which the Average (Best) values in \cref{tab:overall,tab:support} are computed.
Several metrics are non-monotonic across checkpoints.}%
\label{tab:checkpoints}
\begin{tabular}{lrrrrrr}
\toprule
\textbf{Model} & \textbf{Step} & \textbf{FVMD (D)} & \textbf{FVMD (S)} & \textbf{Best-of-$K$} & \textbf{LRTL (ALL)} & \textbf{LRTL (FG)} \\
\midrule
\multirow{5}{*}{S}
& 60000  & 13068.82 & 17512.31 & 204.86 & 9.27 & 7.41 \\
& 70000  & 12237.69 & 16350.72 & 169.90 & 7.54 & 6.26 \\
& 80000  & 12328.60 & 16447.45 & 155.07 & 7.51 & 6.46 \\
& 90000  & 12746.66 & 16941.53 & 141.66 & 7.23 & 6.45 \\
& 100000 & 12349.56 & 16597.45 & 129.77 & 6.91 & 6.30 \\
\midrule
\multirow{4}{*}{S (Dense)}
& 60000  & 11957.53 & 15733.35 & 137.26 & 6.64 & 5.46 \\
& 70000  & 11630.86 & 15452.36 & 138.41 & 6.65 & 5.63 \\
& 80000  & 11591.28 & 15592.10 & 139.02 & 7.24 & 6.03 \\
& 100000 & 12049.17 & 16003.89 & 128.69 & 6.80 & 6.17 \\
\midrule
\multirow{5}{*}{B}
& 60000  & 11812.41 & 15828.66 & 130.98 & 7.27 & 6.50 \\
& 70000  & 11957.17 & 16042.63 & 126.12 & 6.77 & 6.19 \\
& 80000  & 14304.91 & 18627.16 & 129.30 & 6.83 & 5.95 \\
& 90000  & 14615.27 & 18911.21 & 120.19 & 6.37 & 5.82 \\
& 100000 & 14880.14 & 19169.12 & 115.24 & 6.13 & 5.58 \\
\midrule
\multirow{5}{*}{B (Dense)}
& 60000  & 12333.12 & 15958.49 & 120.30 & 6.13 & 5.31 \\
& 70000  & 12738.38 & 16730.78 & 118.76 & 6.35 & 5.66 \\
& 80000  & 13391.11 & 17745.91 & 141.90 & 8.05 & 6.47 \\
& 90000  & 13769.36 & 18143.02 & 131.45 & 6.76 & 5.90 \\
& 100000 & 14096.63 & 18606.89 & 139.77 & 6.72 & 6.00 \\
\bottomrule

\end{tabular}
\end{table}

\FloatBarrier
\section{Implementation details}%
\label{app:impl}

\Cref{tab:model_card} lists the \nickname{} variants, \cref{tab:shared_arch_hyperparams} the architecture hyperparameters shared by all variants, and \cref{tab:hyperparameters} the training and sampling hyperparameters.

\begin{table}[h]
\centering
\small
\caption{\nickname model variants.}%
\label{tab:model_card}
\begin{tabular}{lccccc}
\toprule
\textbf{Model Variant} & \textbf{Enc. Depth} & \textbf{Dec. Depth} & \textbf{Hidden Size ($d_{model}$)} & \textbf{Heads} & \textbf{Head Dim} \\
\midrule
\nickname-XS & 6  & 2 & 192  & 3  & 64 \\
\nickname-S  & 12 & 2 & 384  & 6  & 64 \\
\nickname-B  & 12 & 4 & 768  & 12 & 64 \\
\bottomrule
\end{tabular}
\end{table}
\begin{table}[h]
\centering
\small
\caption{Shared architecture hyperparameters.}%
\label{tab:shared_arch_hyperparams}
\begin{tabularx}{\textwidth}{llX}
\toprule
\textbf{Parameter} & \textbf{Default} & \textbf{Description} \\
\midrule
\texttt{traj\_length} & 24 & Number of frames/timesteps in the trajectory sequence. \\
\texttt{in\_channels} & 2 & Input coordinate dimensions (e.g., $x$ and $y$). \\
\texttt{frame\_feature\_size} & 384 & Dimensionality of the pre-extracted vision features. \\
\texttt{frame\_feature\_patch\_size} & 14 & Spatial patch size used by the vision backbone. \\
\texttt{mlp\_ratio} & 4.0 & Expansion factor for the hidden dimension in the MLP blocks. \\
\texttt{num\_freq\_bands} & 8 & Number of frequency bands for Fourier/Sin-Cos featurization. \\
\texttt{sample\_channels} & 3 & Number of channels to sample from the reference feature map. \\
\texttt{sample\_patch\_size} & 1 & Spatial size of the local patch sampled around each query point coordinate. \\
\texttt{sample\_method} & \texttt{bilinear} & Interpolation method for coordinate-based grid sampling. \\
\texttt{attn\_drop} & 0.0 & Dropout probability for attention weights (applied to middle layers). \\
\texttt{proj\_drop} & 0.0 & Dropout probability for linear projections (applied to middle layers). \\
\bottomrule
\end{tabularx}
\end{table}
\begin{table}[h]
\centering
\caption{Hyperparameters for \nickname.}%
\label{tab:hyperparameters}
\begin{tabular}{@{}lll@{}}
\toprule
\textbf{} & \textbf{Parameter} & \textbf{Value} \\ \midrule
\textbf{Model}
& Stochastic Interpolant Path  & Linear (Rectified-Flow) \\
& Stochastic Interpolant Target & velocity \\
& Network Prediction & velocity \\
& Parameterization & velocities $\velocity_t$ \\
& Feature Extractor & DINOv2 \\
& Feature Extractor Size & Small \\
& Feature Extractor Registers & 4 \\
& Feature Extractor Layers & 3, 6, 9, 12 \\
& Layer-Normalize Features & True \\
& Query Sample Patch Size & 1 \\
\textbf{Training} & Optimizer & AdamW \\
& Learning Rate & $6 \times 10^{-5}$ \\
& Learning Rate Schedule & constant \\
& Weight Decay & 0.0 \\
& Adam Betas & (0.9, 0.999) \\
& EMA Decay & 0.9999 \\
& Global Batch Size & 64 \\
& Total Steps & 100,000 \\
& Global Seed & 42 \\
& Number of Queries & $\operatorname{Uniform}[1,1024]$ \\
& Timestep ($\tau$) Distribution & $\operatorname{LogitNormal}(0,1)$ \\
\midrule
\textbf{Sampling} & SDE Diffusion & SBDM \\
& Solver & Euler-Maruyama \\
& Solver Steps & 1000 \\
& Last Step Type & Mean \\
& Last Step Size & $\frac{1}{1000}$ \\
\midrule
\textbf{Hardware} & Physical Nodes & 1 \\
& GPUs per Node & 2 \\
& GPU Type & NVIDIA GH200 \\
& Workers & 8 \\
& Training Max Walltime & 24 hours \\ \bottomrule
\end{tabular}
\end{table}

\clearpage
\section{Qualitative results on Kubric-A}%
\label{app:qualitative-kubric}
\begin{figure}[!b]
\centering
\includegraphics[width=0.95\textwidth]{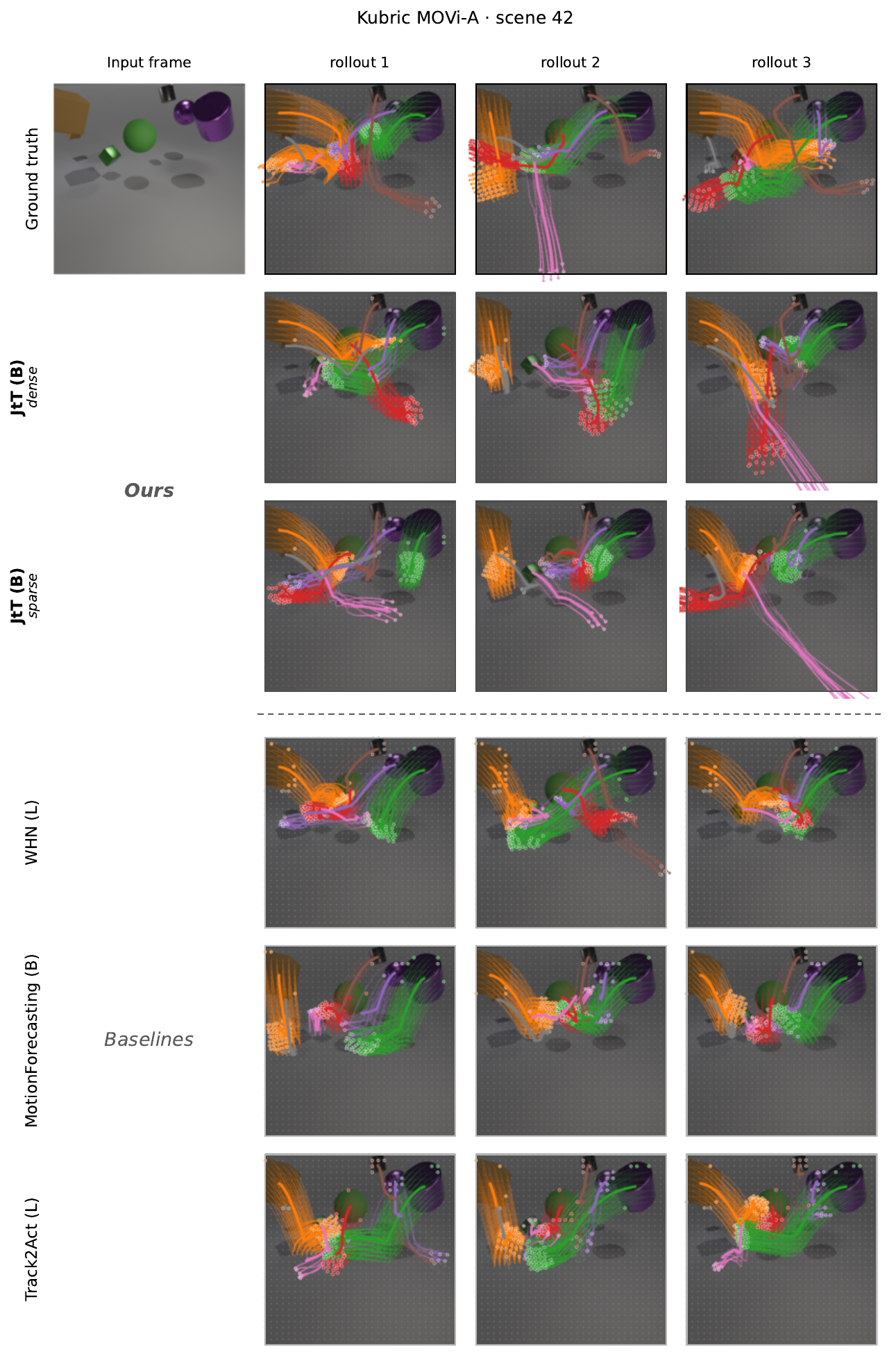}
\caption{\textbf{Comparison on the Kubric benchmark of \WHN{}~\citep{boduljak2026whn}.}
Scene 42.}%
\label{fig:qualitative-kubric-000042}
\end{figure}

\begin{figure}[p]
\centering
\includegraphics[width=\textwidth]{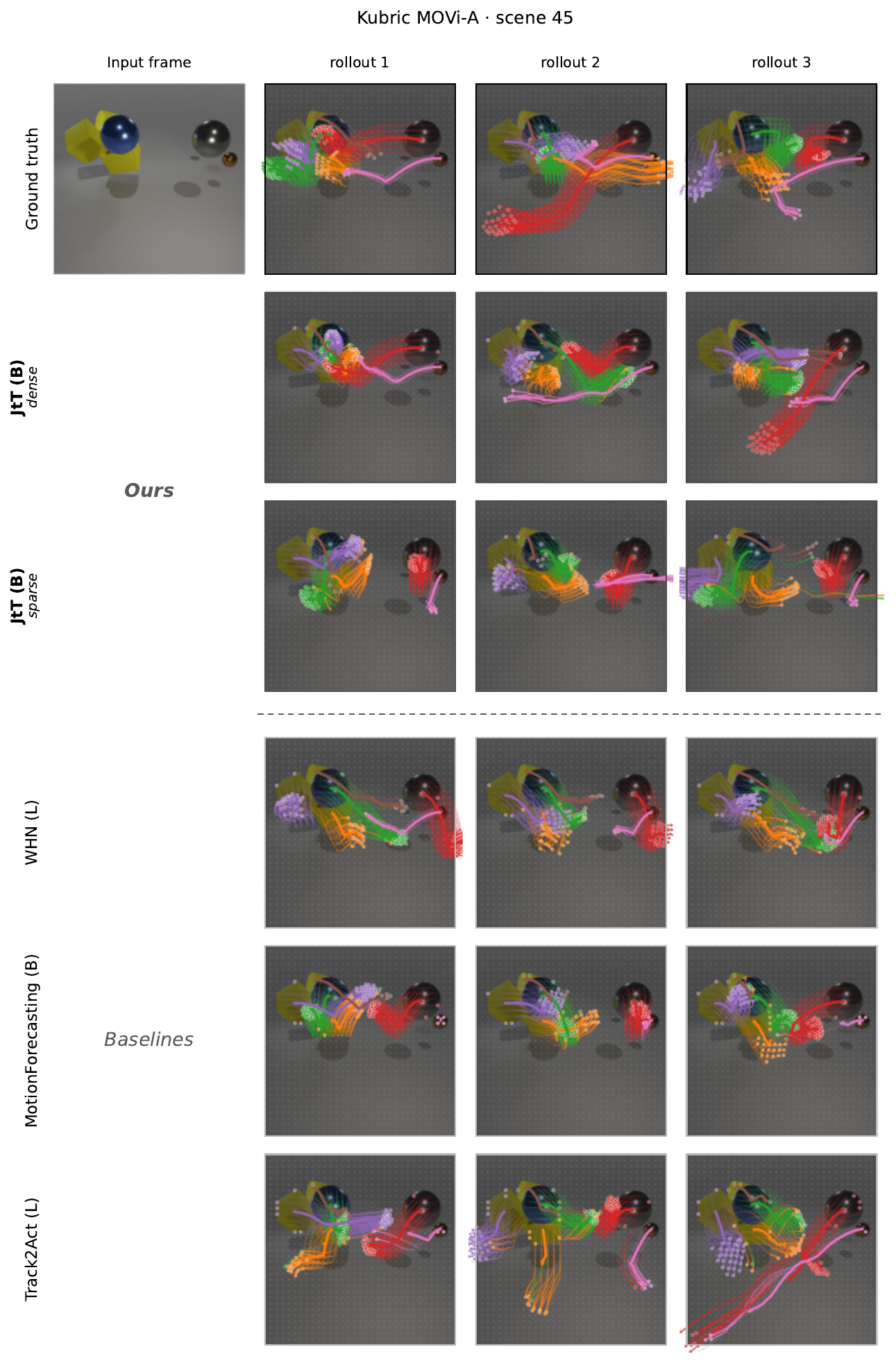}
\caption{\textbf{Comparison on the Kubric benchmark of \WHN{}~\citep{boduljak2026whn}.}
Scene 45.}%
\label{fig:qualitative-kubric-000045}
\end{figure}

\begin{figure}[p]
\centering
\includegraphics[width=\textwidth]{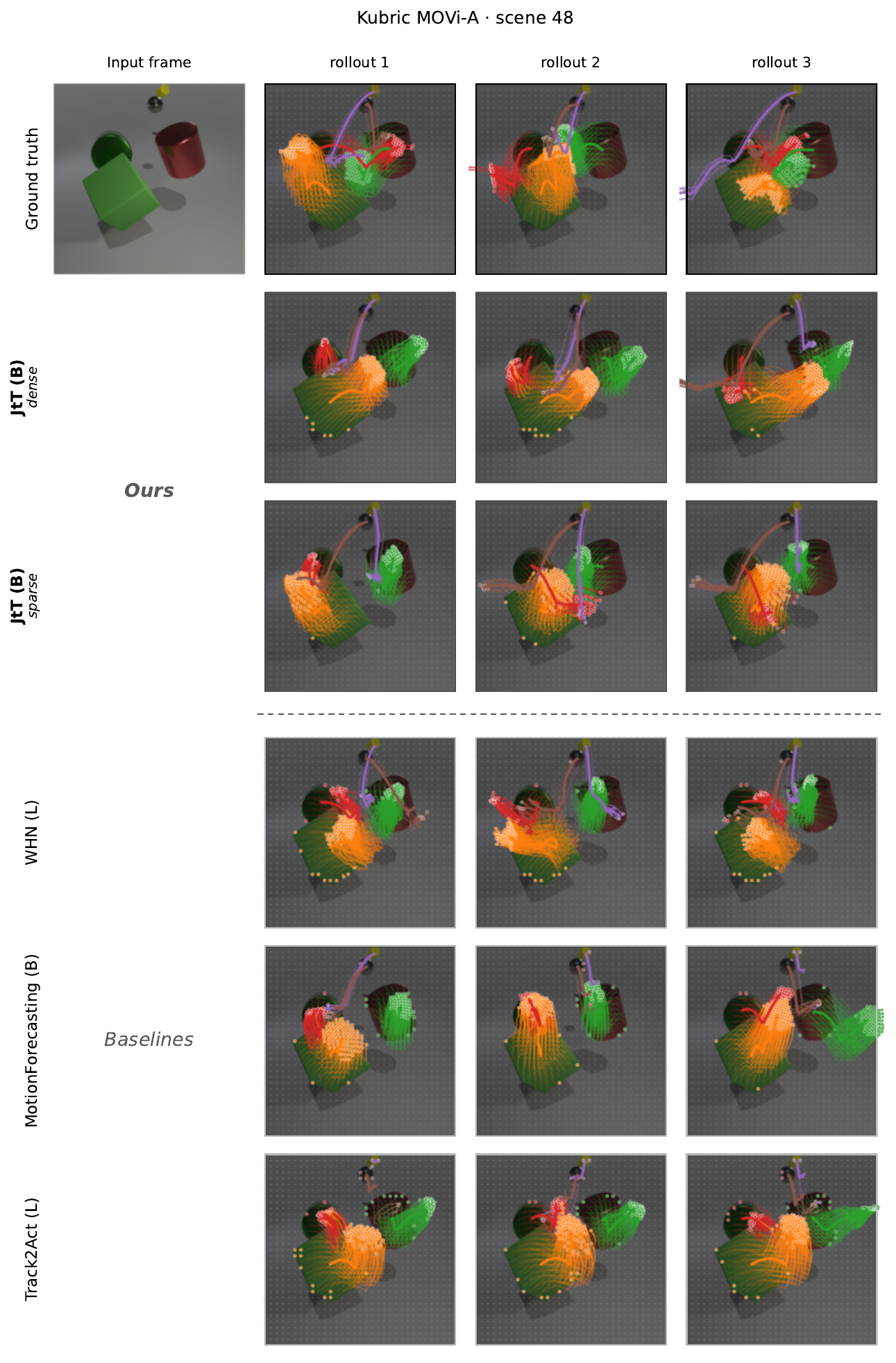}
\caption{\textbf{Comparison on the Kubric benchmark of \WHN{}~\citep{boduljak2026whn}.}
Scene 48.}%
\label{fig:qualitative-kubric-000048}
\end{figure}

\begin{figure}[p]
\centering
\includegraphics[width=\textwidth]{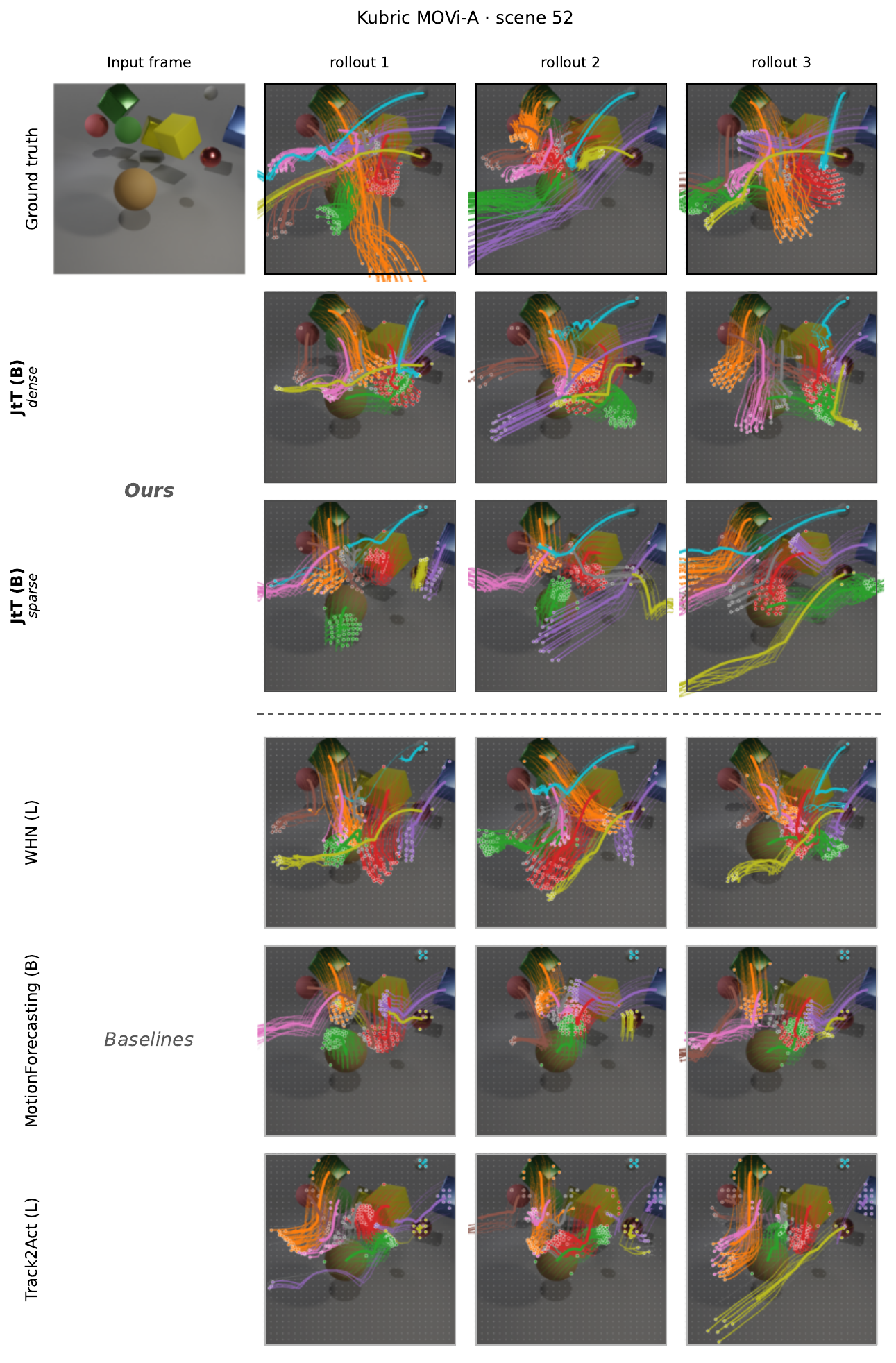}
\caption{\textbf{Comparison on the Kubric benchmark of \WHN{}~\citep{boduljak2026whn}.}
Scene 52.}%
\label{fig:qualitative-kubric-000052}
\end{figure}

\begin{figure}[p]
\centering
\includegraphics[width=\textwidth]{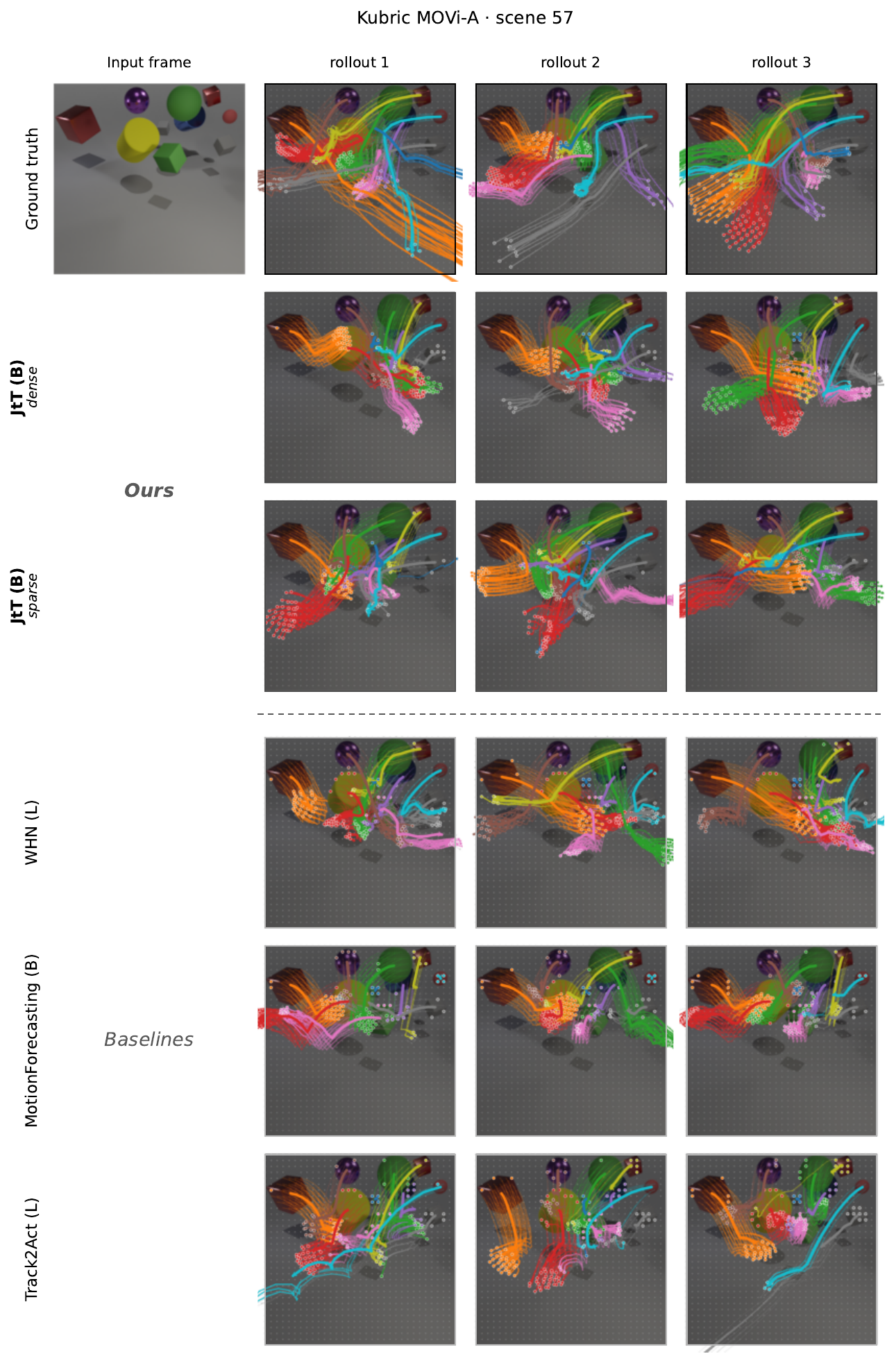}
\caption{\textbf{Comparison on the Kubric benchmark of \WHN{}~\citep{boduljak2026whn}.}
Scene 57.}%
\label{fig:qualitative-kubric-000057}
\end{figure}

\clearpage
\section{Qualitative results on Physics 101}%
\begin{figure}[!ht]
\centering
\includegraphics[width=\textwidth]{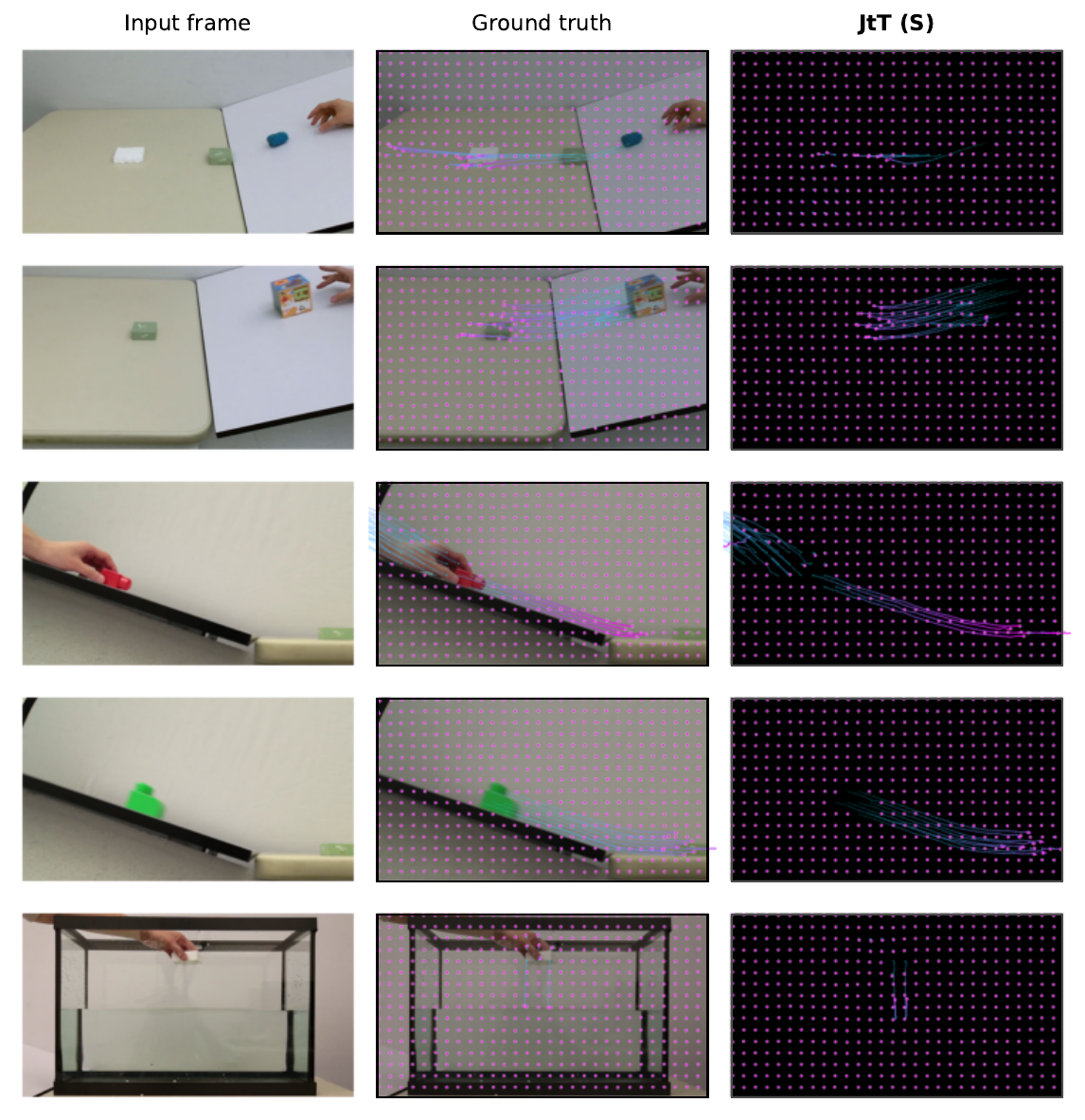}
\caption{\textbf{Qualitative results on Physics 101}~\citep{physics101}, following the training and evaluation setup of \WHN{}~\citep{boduljak2026whn}.}%
\label{fig:qualitative-physics101}
\end{figure}

\begin{figure}[!ht]
\centering
\includegraphics[width=\textwidth]{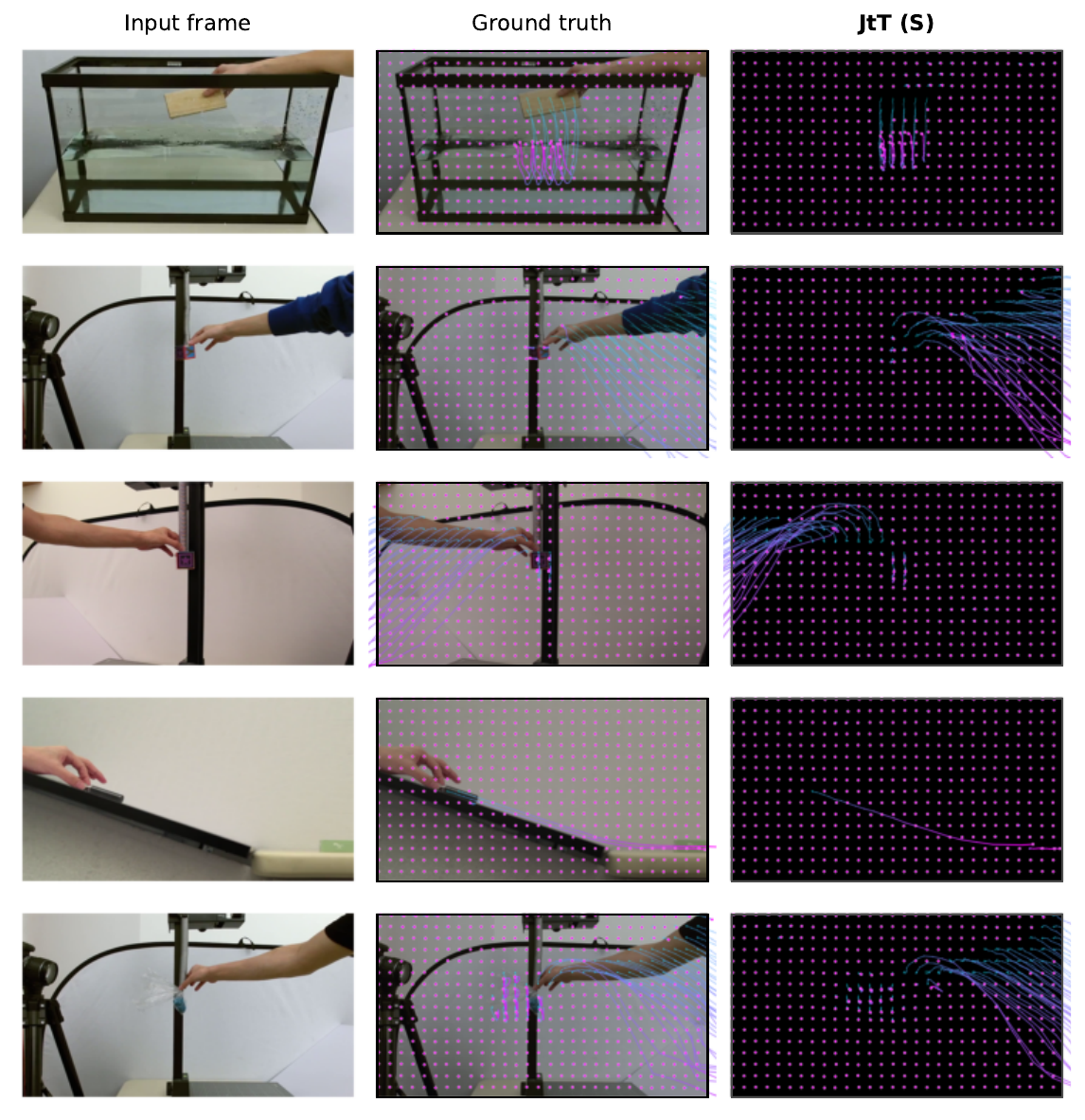}
\caption{\textbf{More qualitative results on Physics 101}~\citep{physics101}, following the training and evaluation setup of \WHN{}~\citep{boduljak2026whn}.}%
\label{fig:qualitative-physics101-2}
\end{figure}

\begin{figure}[!ht]
\centering
\includegraphics[width=\textwidth]{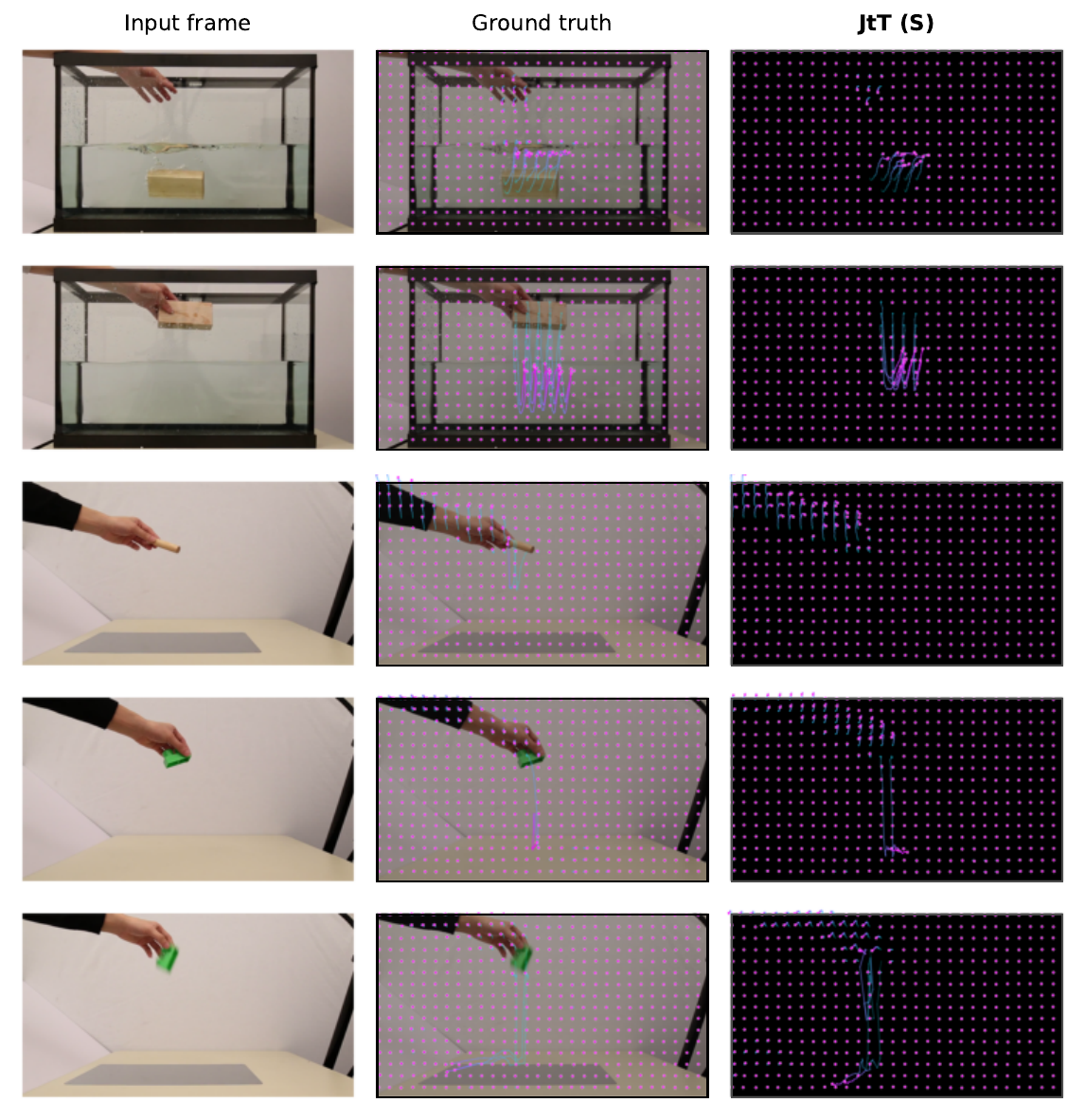}
\caption{\textbf{More qualitative results on Physics 101}~\citep{physics101}, following the training and evaluation setup of \WHN{}~\citep{boduljak2026whn}.}%
\label{fig:qualitative-physics101-3}
\end{figure}

\begin{figure}[!ht]
\centering
\includegraphics[width=\textwidth]{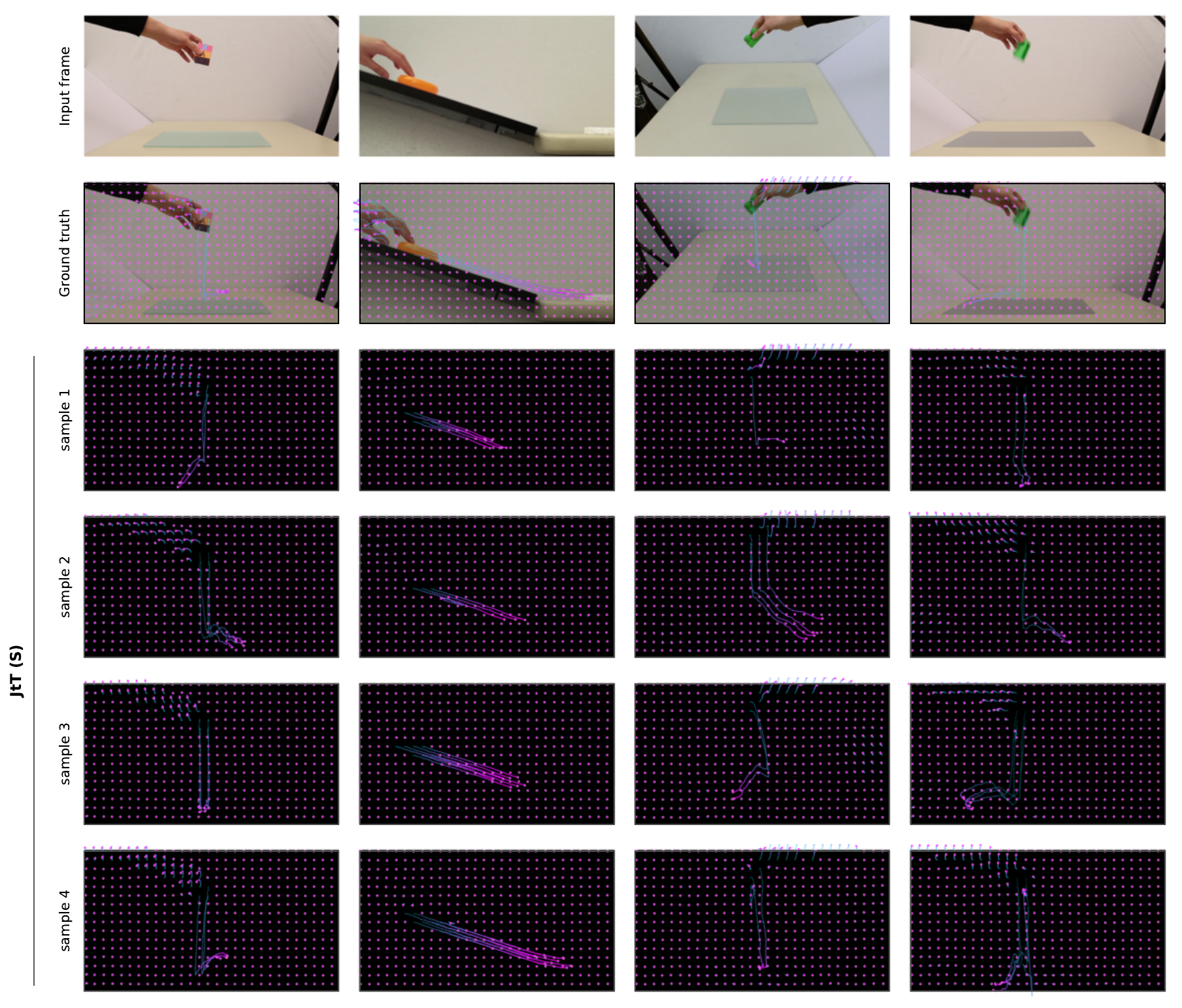}
\caption{\textbf{Sample diversity on Physics 101}~\citep{physics101}.
Four samples from \nickname{} per input frame, following the training and evaluation setup of \WHN{}~\citep{boduljak2026whn}.}%
\label{fig:qualitative-physics101-diversity}
\end{figure}

\label{app:qualitative-physics101}
Following \WHN{}~\citep{boduljak2026whn}, we also train \nickname{} on real videos from Physics 101~\citep{physics101}, which records simple physical experiments, and show samples on held-out sequences.
As for Cityscapes (\cref{app:qualitative-cityscapes}), real videos provide a single observed future per image and no object-level rigidity ground truth, so a rigorous evaluation is not possible and we report qualitative results only (\cref{fig:qualitative-physics101,fig:qualitative-physics101-2,fig:qualitative-physics101-3}).
\nickname{} generates rigid motion for rigid bodies, such as objects sliding down a ramp or falling onto a table.
It also captures the interaction between rigid and non-rigid bodies, such as a hand releasing an object or a block entering water, and propagates forces through contact, such as a sliding object pushing another one on collision.
Finally, it reflects different kinds of forces acting on rigid bodies, such as gravity and buoyancy on a wooden block dropped into water.
\nickname{} also retains its ability to generate diverse and plausible futures on real data: for the same input frame, samples differ in how objects slide, fall, and bounce, while remaining consistent with the ground truth (\cref{fig:qualitative-physics101-diversity}).

\clearpage
\section{Qualitative results on Cityscapes}%
\begin{figure}[!ht]
\centering
\includegraphics[width=\textwidth]{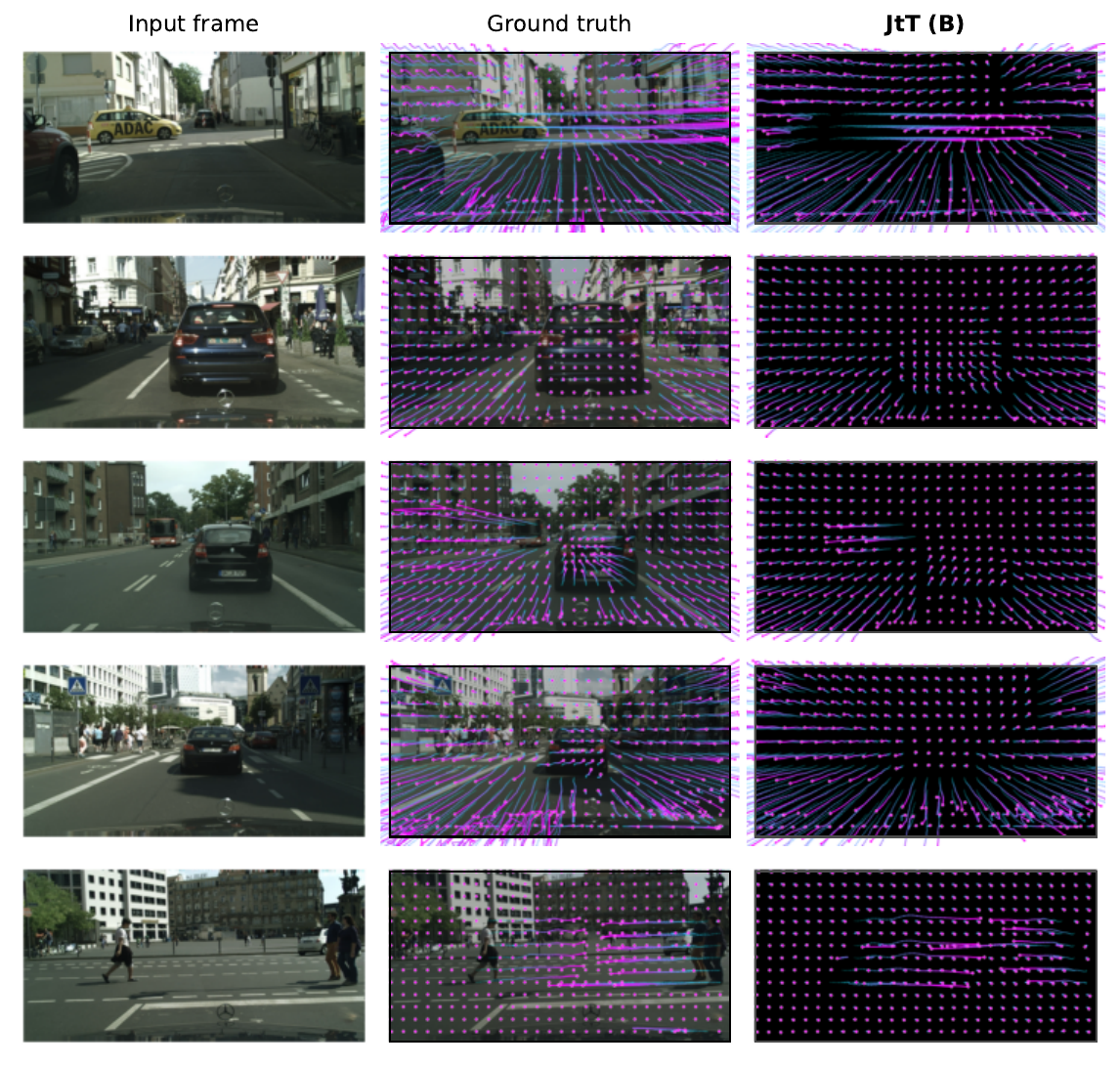}
\caption{\textbf{Qualitative results on Cityscapes}~\citep{cordts16the-cityscapes}, using the training and evaluation split of \WHN{}~\citep{boduljak2026whn}.}%
\label{fig:qualitative-cityscapes}
\end{figure}

\begin{figure}[!ht]
\centering
\includegraphics[width=\textwidth]{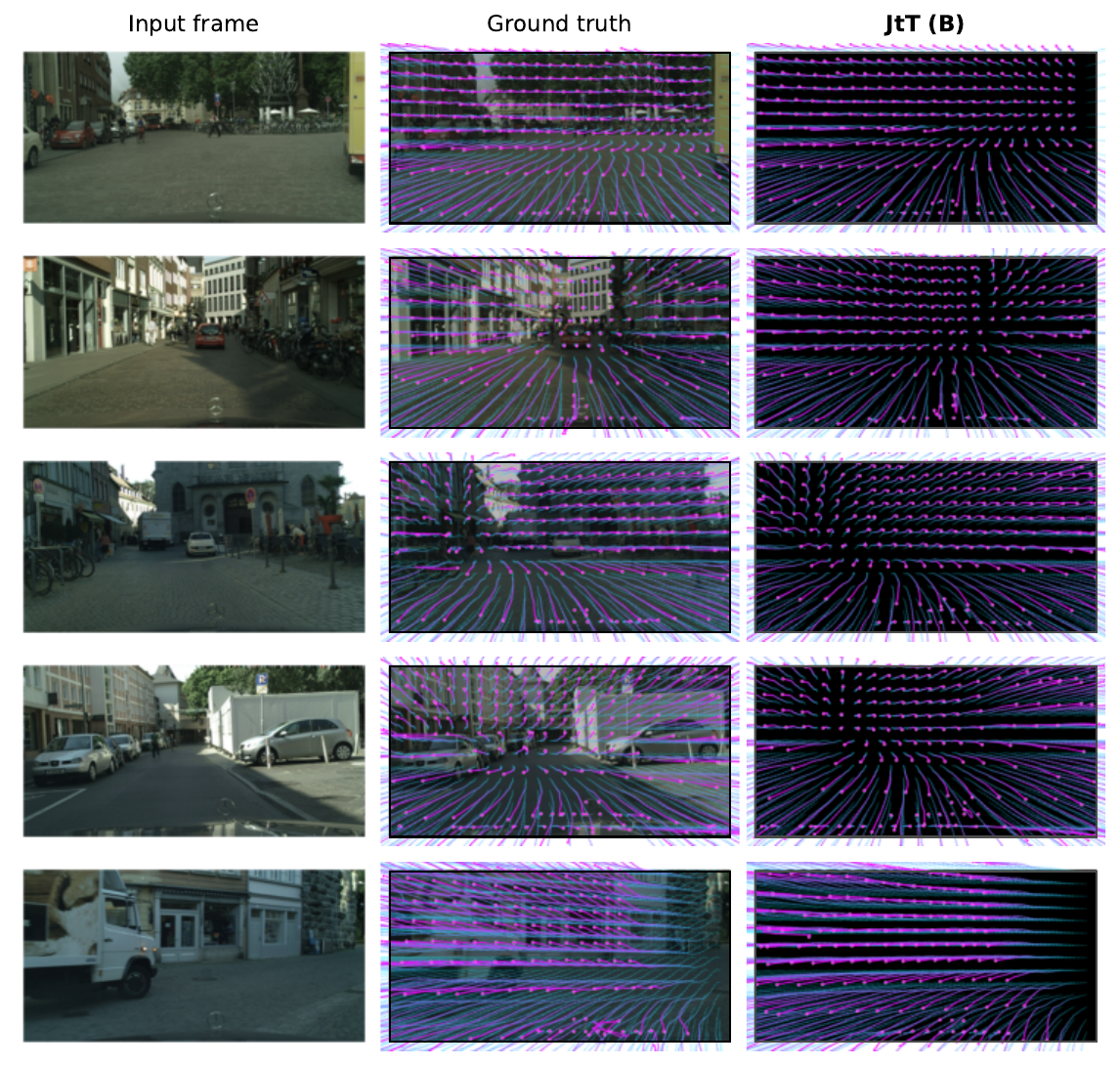}
\caption{\textbf{More qualitative results on Cityscapes}~\citep{cordts16the-cityscapes}, using the training and evaluation split of \WHN{}~\citep{boduljak2026whn}.}%
\label{fig:qualitative-cityscapes-2}
\end{figure}

\begin{figure}[!ht]
\centering
\includegraphics[width=\textwidth]{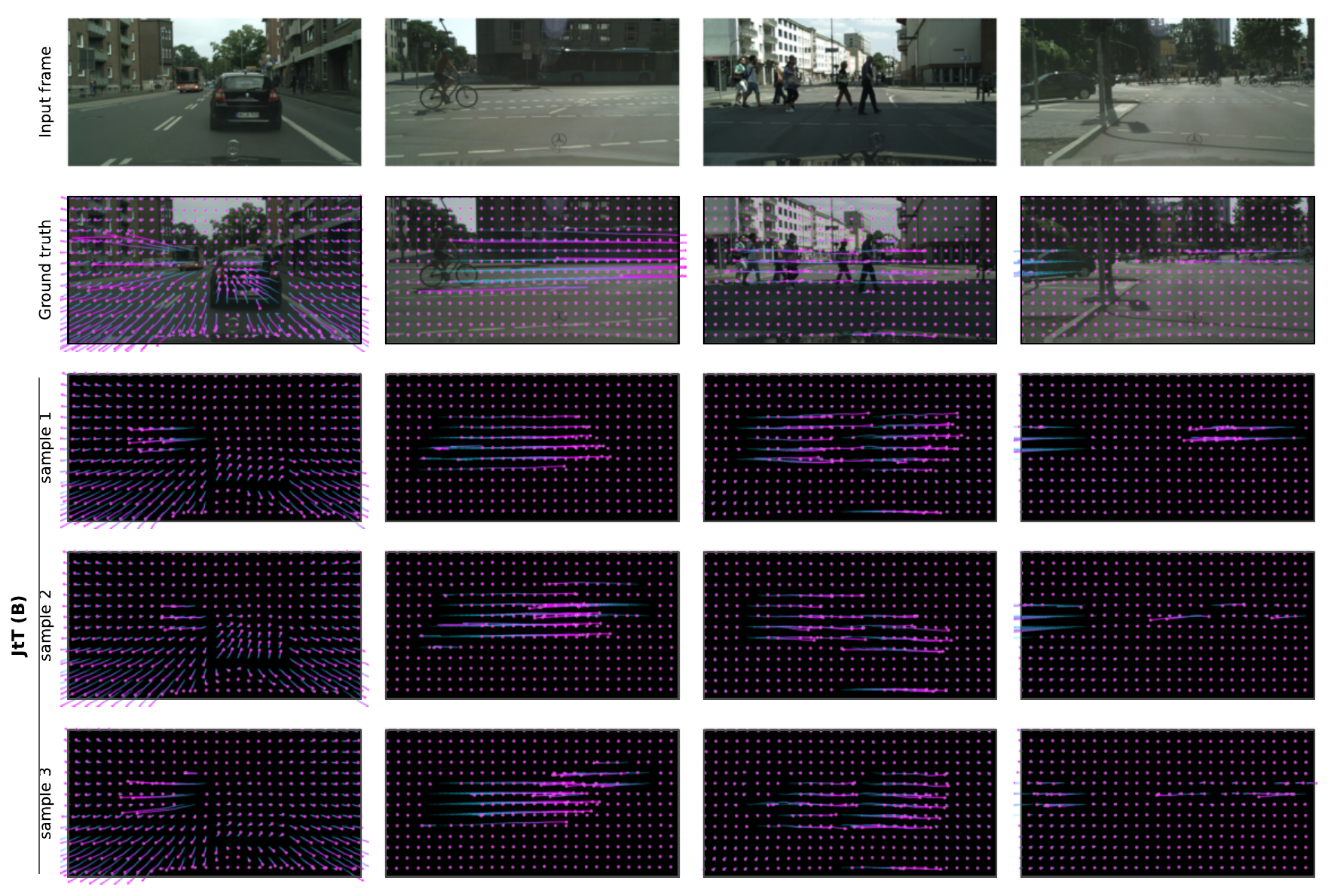}
\caption{\textbf{Sample diversity on Cityscapes}~\citep{cordts16the-cityscapes}.
Three samples from \nickname{} per input frame, using the training and evaluation split of \WHN{}~\citep{boduljak2026whn}.}%
\label{fig:qualitative-cityscapes-diversity}
\end{figure}

\label{app:qualitative-cityscapes}
To test whether \nickname{} transfers to real data, we train it on Cityscapes~\citep{cordts16the-cityscapes}, using the training and evaluation data of \WHN{}~\citep{boduljak2026whn}.
This analysis is qualitative: real videos provide a single observed future per image and no object-level rigidity ground truth, so the rigorous evaluation of our synthetic experiments (\cref{sec:experiments}) is not possible here.
The training set has 2,975 sequences, the same order of magnitude as Kubric-A.
We train for the same number of steps as on Kubric-A, with identical hyperparameters (\cref{tab:hyperparameters}), and stop training once we observe overfitting on the held-out set.
\Cref{fig:qualitative-cityscapes,fig:qualitative-cityscapes-2} show samples.
Qualitatively, \nickname{} preserves the rigidity of objects that should move rigidly, such as cars, and handles egocentric camera motion well.
It also generalizes to some extent to deformable objects such as pedestrians: their overall motion is plausible, but less precise than for rigid objects.
We also verify that \nickname{} samples diverse futures consistent with the ground truth (\cref{fig:qualitative-cityscapes-diversity}).
It does, and the samples are reasonable, although less diverse than on Kubric-A, likely because it is trained on about $5{\times}$ less data.

\end{document}